\documentclass[times,twocolumn,final]{elsarticle}
\usepackage{medima_arxiv}
\usepackage{framed,multirow}

\usepackage{amssymb}
\usepackage{latexsym}

\usepackage{url}
\usepackage{xcolor}

\usepackage[colorlinks=true,linkcolor=blue,urlcolor=blue,citecolor=blue]{hyperref}

\definecolor{newcolor}{rgb}{.8,.349,.1}

\usepackage{graphicx}
\usepackage{scalerel,stackengine}

\usepackage{epsfig}
\usepackage{multirow}
\usepackage{amsmath}
\usepackage{amssymb}
\usepackage{booktabs}
\usepackage{ dsfont }
\usepackage{ cite }
\usepackage{float}
\usepackage{color}
\usepackage{xcolor}
\usepackage{soul}
\usepackage{float}
\usepackage{url}
\usepackage{rotate}
\usepackage{tikz}
\usepackage{makecell}
\usepackage{tabularx}
\usepackage{algorithm}
\usepackage{ifthen}
\usepackage{tablefootnote}
\usepackage{textcomp}
\usepackage{subfigure}

\usepackage[normalem]{ulem}

\usepackage{algorithmicx}
\usepackage[noend]{algpseudocode}
\usepackage{mathtools}

\DeclareMathOperator*{\argmax}{arg\,max}
\DeclareMathOperator*{\argmin}{arg\,min}

\definecolor{alizarin}{rgb}{0.82, 0.1, 0.26}
\definecolor{aliceblue}{rgb}{0.94, 0.97, 1.0}
\definecolor{amber}{rgb}{1.0, 0.75, 0.0}
\definecolor{amethyst}{rgb}{0.6, 0.4, 0.8}

\definecolor{r1}{rgb}{0.0, 0.0, 0.0}
\definecolor{r2}{rgb}{0.0, 0.0, 0.0}

\newcommand{\blueb}[1]{\textcolor{blue}{\textbf{#1}}}

\makeatletter

\newcolumntype{Y}{>{\centering\arraybackslash}X}
\algnewcommand{\nComment}[1]{\Statex \Comment{#1}}

\algdef{SE}[SUBALG]{Indent}{EndIndent}{}{\algorithmicend\ }%
\algtext*{Indent}
\algtext*{EndIndent}
\newcounter{phase}[algorithm]
\newlength{\phaserulewidth}
\newcommand{\setphaserulewidth}{\setlength{\phaserulewidth}}

\makeatother

\setphaserulewidth{.7pt}

\journal{Medical Image Analysis}

\begin{document}

\newcolumntype{C}[1]{>{\centering\arraybackslash\hspace{0pt}}p{#1}}

\verso{Yuanhong Chen \textit{et~al.}}

\begin{frontmatter}

\title{
BRAIxDet: Learning to Detect Malignant Breast Lesion with Incomplete Annotations
}%
% \tnotetext[tnote1]{This is an example for title footnote coding.}

\author[1]{Yuanhong \snm{Chen}}
\author[1]{Yuyuan \snm{Liu}}
\author[1]{Chong \snm{Wang}\corref{cor1}}
\cortext[cor1]{Corresponding author.}
\ead{chong.wang@adelaide.edu.au}
\author[2]{Michael \snm{Elliott}}
\author[2]{Chun Fung \snm{Kwok}}
\author[2]{Carlos \snm{Peña-Solorzano}} 
\author[1]{Yu \snm{Tian}}
\author[1]{Fengbei \snm{Liu}}
\author[3]{Helen \snm{Frazer}}
\author[2,4]{Davis J. \snm{McCarthy}}
\author[1]{Gustavo \snm{Carneiro}}

% \author[1]{Given-name1 \snm{Surname1}\corref{cor1}}
% \cortext[cor1]{Corresponding author: 
%   Tel.: +0-000-000-0000;  
%   fax: +0-000-000-0000;}
% \author[1]{Given-name2 \snm{Surname2}\fnref{fn1}}
% \fntext[fn1]{This is author footnote for second author.}
% \author[2]{Given-name3 \snm{Surname3}}
% %% Third author's email
% \ead{author3@author.com}
% \author[2]{Given-name4 \snm{Surname4}}

\address[1]{Australian Institute for Machine Learning, The University of Adelaide, Adelaide, Australia}
\address[2]{Bioinformatics and Cellular Genomics, St Vincent's Institute of Medical Research, Melbourne, Australia}
\address[3]{St Vincent's Hospital Melbourne, Melbourne, Australia}
\address[4]{Melbourne Integrative Genomics, The University of Melbourne, Melbourne, Australia}

\received{1 May 2013}
\finalform{10 May 2013}
\accepted{13 May 2013}
\availableonline{15 May 2013}
% \communicated{S. Sarkar}

\begin{abstract}
%%%
Methods to detect malignant lesions from screening mammograms are usually trained with fully annotated datasets, where images are labelled with the localisation and classification of cancerous lesions.
However, real-world screening mammogram datasets commonly have a subset that is fully annotated and another subset that is weakly annotated with just the global classification (i.e., without lesion localisation).
Given the large size of such datasets, researchers usually face a dilemma with the weakly annotated subset: to not use it or to fully annotate it.  
The first option will reduce detection accuracy because it does not use the whole dataset, 
and the second option is too expensive given that the annotation needs to be done by expert radiologists.
In this paper, we propose a middle-ground solution for the dilemma, which is to formulate the training as a weakly- and semi-supervised learning problem that we refer to as malignant breast lesion detection with incomplete annotations.
To address this problem, our new method comprises two stages, namely: 1) pre-training a multi-view mammogram classifier with weak supervision from the whole dataset, and 2) extending the trained classifier to become a multi-view detector that is trained with semi-supervised student-teacher learning, where the training set contains fully and weakly-annotated mammograms.
We provide extensive detection results on two real-world screening mammogram datasets containing incomplete annotations and show that our proposed approach achieves state-of-the-art results in the detection of malignant breast lesions with incomplete annotations.
%%%%
\end{abstract}

\begin{keyword}
%% MSC codes here, in the form: \MSC code \sep code
%% or \MSC[2008] code \sep code (2000 is the default)
% \MSC 41A05\sep 41A10\sep 65D05\sep 65D17
%% Keywords
\KWD 
\newline
Deep learning \\
Multi-view learning \\
Breast cancer screening \\
Incomplete annotations \\
Malignant lesion detection \\
Student-teacher Learning \\
\end{keyword}

\end{frontmatter}

%\linenumbers

%% main text
% \input{sections/1_intro_literature}
% \input{sections/2_method}
% \input{sections/3_experiments}

% \input{major_revision_sections/1_intro_literature}
% \input{major_revision_sections/2_method}
% \input{major_revision_sections/3_experiments}

\section{Introduction}

Breast cancer is the most commonly diagnosed cancer worldwide and the leading cause of cancer-related death in women~\citep{sung2021global}. 
One of the most effective ways to increase the survival rate relies on the  early detection of breast cancer~\citep{lauby2015breast} using
screening mammograms~\citep{selvi2014breast}.
The analysis of screening mammograms is generally done manually by radiologists, with the eventual help of Computer-Aided Diagnosis (CAD) tools to assist with the detection and classification of breast lesions~\citep{hadjiiski2006advances,shen2019deep}. However, CAD tools have shown inconsistent results in clinical settings. 
Some research groups~\citep{brem2003improvement,hupse2009use, hupse2013standalone} have shown promising results, where radiologists benefited from the use of CAD systems with an improved detection sensitivity. However, other studies~\citep{lehman2015diagnostic,fenton2007influence,fenton2011effectiveness} do not support the reimbursement of the costs associated with the use of CAD systems, estimated to be around \$400 million a year~\citep{lehman2015diagnostic}, since experimental results show no benefits in terms of detection rate. 

% DNN
The use of deep learning~\citep{lecun2015deep} for the analysis of screening mammograms is a promising approach to improve the accuracy of CAD tools, given its outstanding performance in other learning tasks~\citep{he2016deep,ren2015faster}.
For example, \citet{shen2019globally,shen2021interpretable} proposed a classifier that relies on concatenated features from a global network (using the whole image) and a local network (using image patches). To leverage the cross-view information of mammograms, \citet{carneiro2017automated} explored different feature fusion strategies to integrate the knowledge from multiple mammographic views to enhance classification performance.
However, these studies that focus on classification without producing reliable lesion detection results may not be useful in clinical practice. % To enable the localize of malignant region, 
Lesion detection from mammograms has been studied by~\citet{ribli2018detecting} and~\citet{dhungel2015automated}, who developed CAD systems based on modern visual object detection methods.
\citet{ma2021cross} implemented a relation network~\citep{hu2018relation} to learn the inter-relationships between the region proposals from ipsi-lateral mammographic views. \citet{yang2020momminet,yang2021momminetv2} proposed a system that focused on the detection and classification of masses from mammograms by exploring complementary information from ipsilateral and bilateral mammographic views. 

\begin{figure}[t]
    \centering
    \includegraphics[width=1.0\linewidth]{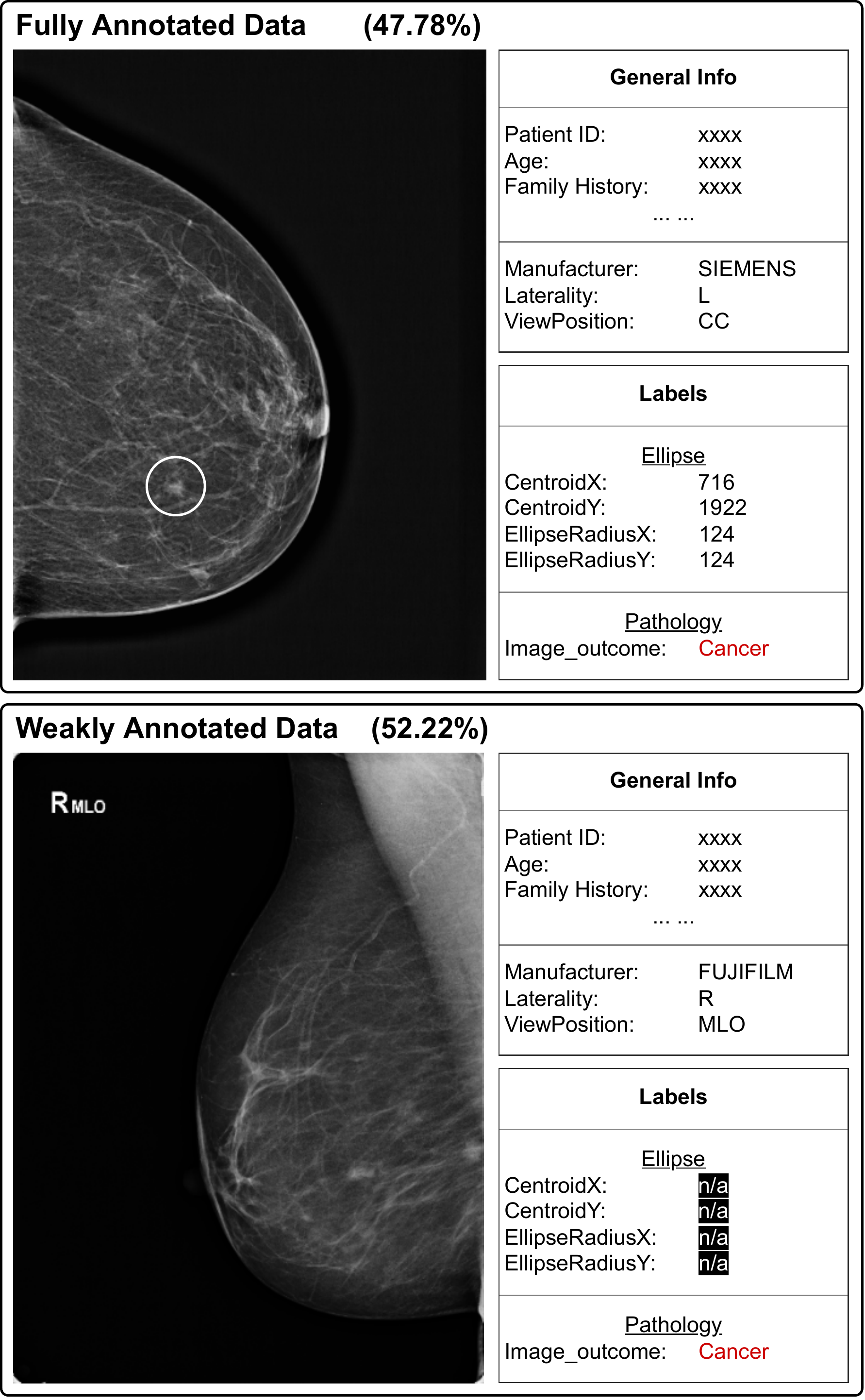}
    \caption{The data structure of our ADMANI dataset, showing 
    fully (top, representing 47.78\% of the data) and weakly (bottom, representing the remaining 52.22\% of the data) annotated mammograms.}
    \label{fig:weakly_semi_setup}
\end{figure}

Although the mammographic lesion detection methods above show encouraging performance, they are trained on fully annotated datasets, where images are labelled with the classification and localisation of cancerous lesions.
Such setup is uncommon in large-scale real-world screening mammogram datasets that usually
have a subset of the images  annotated with classification and localisation of lesions, and the other subset annotated with only the global classification.  
For example, our own Annotated Digital Mammograms and Associated Non-Image (ADMANI) dataset, which has been collected from the BreastScreen Victoria (Australia) population screening program, has a total of 4,084,098 images, where only 47.78\% of the 21,399 cancer cases are annotated with classification and localisation of lesions, and the rest 52.22\% of the cases are weakly annotated, as displayed in Figure~\ref{fig:weakly_semi_setup}.
Given the large size of such datasets, it is rather disadvantageous not to use these weakly annotated images for training, but at the same time, it is too expensive to have them annotated by radiologists.
Therefore, in this paper we propose a middle-ground solution, which is to formulate the training as a weakly- and semi-supervised learning problem that is referred to as malignant breast lesion detection with incomplete annotations.

To address the incomplete annotations learning problem described above, we introduce a novel multi-view lesion detection method, called BRAIxDet. 
BRAIxDet is trained using a two-stage learning strategy that utilises all training samples from the dataset. In the first stage, we pre-train our previously proposed multi-view classification network BRAIxMVCCL~\citep{chen2022multi} on the whole weakly-supervised training set, where images are annotated only with their global labels.

This pre-training enables BRAIxMVCCL to produce a strong feature extractor and a post-hoc explanation based on Grad-CAM~\citep{selvaraju2017grad}.
In the second stage, we transform BRAIxMVCCL into the detector BRAIxDet which is trained using a student-teacher semi-supervised learning mechanism. 
More specifically, the student is trained from fully- and weakly-supervised subsets, where the weakly-supervised images have their lesions pseudo-labelled by the teacher's detections and BRAIxMVCCL's Grad-CAM results.
Furthermore, the teacher is trained from the exponential moving average (EMA) of the student's parameters, where
the batch normalization (BN) statistic used in the EMA is frozen after pre-training to alleviate the issues related to the dependency on the samples used to train the student, and the mismatch of model parameters between student and teacher.
The major contributions of this paper are summarised as follows:
\begin{itemize}
    \item \textcolor{r1}{We explore \textbf{a new experimental setting for the detection of malignant breast lesions} using large-scale real-world screening mammogram datasets that have incomplete annotations, with a subset of images annotated with the classification and localisation of lesions and a subset of images annotated only with image classification, thus forming a weakly- and semi-supervised learning problem.}
    \item \textcolor{r1}{We propose a \textbf{new two-stage training method to handle the incomplete annotations} introduced above}.
    The first stage uses the images annotated with their classification labels to enable a weakly-supervised pre-training of our previously proposed multi-view classification model BRAIxMVCCL~\citep{chen2022multi}.
    The second stage transforms the multi-view classifier BRAIxMVCCL into the detector BRAIxDet, which is trained with a student-teacher semi-supervised learning approach that uses both the weakly- and the fully-supervised subsets.
    \item \textcolor{r1}{We also propose innovations to the student-teacher semi-supervised detection learning, where the student is trained using the teacher's detections and BRAIxMVCCL's GradCAM~\citep{selvaraju2017grad} outputs to estimate lesion localisation for the weakly-labelled data;} while the teacher is trained with the temporal ensembling of student's parameters provided by EMA, with the BN parameters frozen after pre-training to mitigate the problems introduced by the dependency on the student training samples, and the mismatch between student and teacher model parameters.
    \item \textcolor{r1}{We provide extensive experiments on two real-world br-east cancer screening mammogram datasets containing incomplete annotations.}
\end{itemize}
Our proposed BRAIxDet model achieves state-of-the-art (S OTA) performance on both datasets in terms of lesion detection measures.

\section{Related work}

\subsection{Lesion detection in mammograms}
% Object detection
Object detection is a fundamental task in computer vision~\citep{liu2020deep}. The existing detection methods can be categorized into two groups:
1) two-stage methods~\citep{girshick2015fast,ren2015faster} that first generate region proposals based on local objectiveness, 
and then classify and refine the detection of these region proposals; and 2) single-stage methods~\citep{redmon2016you, carion2020end} that directly output bounding box predictions and corresponding labels. 
In general, the two-stage methods are preferred for non-real-time medical image analysis applications, such as lesion detection from mammograms~\citep{ma2021cross,yang2020momminet, yang2021momminetv2}, since they usually provide more accurate detection performance.
On the other hand, single-stage methods tend to be less accurate but faster, which is more suitable for real-time applications, such as assisted intervention~\citep{butler2022defense}. Similar to previous papers that propose lesion detection methods from mammograms~\citep{ma2021cross,yang2020momminet, yang2021momminetv2}, we adopt the two-stage method Faster-RCNN~\citep{ren2015faster} as our backbone detector since the detection accuracy is more important than speed in our clinical setting.

% Mass detection
A standard screening mammogram exam contains two ipsilateral projection views of each breast, namely bilateral craniocaudal (CC) and mediolateral oblique (MLO), where radiologists analyse both views simultaneously by searching for global architectural distortions and local cancerous lesions (e.g., masses and calcifications). 
Recently, fully-supervised learning methods (i.e., methods containing training sets with complete lesion localisation and classification annotations) have shown promising results for mass detection from single-view~\citep{ribli2018detecting,dhungel2015automated} and multi-view mammograms~\citep{ma2021cross,liu2020cross,yang2020momminet,yang2021momminetv2}. However, these methods only focus on the detection and classification of masses (i.e., not all cancerous lesions), limiting their 
scope and application. 
Similarly to previous multi-view mammogram analysis methods~\citep{ma2021cross,liu2020cross,yang2020momminet,yang2021momminetv2},
BRAIxMVCCL is designed to learn complementary information from ipsilateral mammogram views (CC and MLO). 
Unlike previous methods that depend on the ill-defined registration between the two views (e.g., based on nipple alignment~\citep{yang2020momminet,yang2021momminetv2}, or geometric features ~\citep{ma2021cross,liu2019unilateral}), BRAIxMVCCL realises multi-view information integration via implicit reasoning with a local co-occurrence module. 
Also, BRAIxMVCCL explicitly considers the global consistency between ipsilateral views to enhance the representation ability. 
Furthermore, BRAIxMVCCL is trained with an end-to-end learning method that is simpler and more efficient than the multi-stage learning algorithms to train other multi-view approaches~\citep{yang2020momminet,yang2021momminetv2}.
% Furthermore
Nevertheless, real-world screening mammogram datasets tend not to exclusively contain fully annotated images -- instead, these datasets usually have a subset that is annotated with lesion localisation and classification, and another subset that is weakly annotated only with the global image classification labels.
Instead of discarding the weakly annotated images or hiring a radiologist to annotate these images, we propose a new weakly- and semi-supervised learning approach that can use both subsets of the dataset, without adding new detection labels.

\subsection{Weakly-supervised object detection}

% Weakly
Weakly supervised disease localisation (WSDL) is a challenging learning problem that consists of training a model to localise a disease in a medical image, even though the training set contains only global image classification labels without any disease localisation labels.
WSDL approaches can be categorized into class activation map (CAM) methods~\citep{zhou2016learning}, multiple instance learning (MIL) methods~\citep{oquab2015object}, and
prototype-based methods~\citep{chen2019looks}.
% Discriminative region problem

\textbf{CAM methods} leverage the gradients of the target disease at the last convolution layer of the classifier to produce a coarse heatmap.
This heatmap highlights areas that contribute most to the classification of the disease in a post-hoc manner. 
For example, \citet{rajpurkar2017chexnet} adopted GradCAM~\citep{selvaraju2017grad} to achieve WSDL in chest X-rays. \citet{lei2020shape} extended the original CAM~\citep{zhou2016learning} to learn the importance of individual feature maps at the last convolution layer to capture fine-grained lung nodule shape and margin for lung nodule classification.
\textbf{MIL-based methods} regard each image as a bag of instances and encourage the prediction score for positive bags to be larger than for the negative ones. To extract region proposals included in each bag, early methods~\citep{oquab2015object} adopted max pooling to concentrate on the most discriminative regions. 
To avoid the selection of imprecise region proposals, Noisy-OR~\citep{wang2017chestx}, softmax~\citep{seibold2020self}, and Log-Sum-Exp~\citep{wang2017chestx, yao2018weakly} pooling methods have been used to maintain the relative importance between instance-level predictions and encourage more instances to affect the bag-level loss. 
\textbf{Prototype-based methods} aim to learn class-specific prototypes, represented by learned local features associated with each class, where classification and detection are achieved by assessing the similarity between image and prototype features, producing a detection map per prototype~\citep{chen2019looks}.
The major weakness of WSDL methods is that they are trained only with image-level labels, so the model can only focus on the most discriminative features of the image, which generally produce too many false positive lesion detections.

\subsection{Semi-supervised object detection}
\label{sec:semi-supervised_object_detection}

% Semi
Large-scale real-world screening mammogram datasets usually contain a subset of fully-annotated images, where all lesions have localisation and classification labels, and another subset of images with incomplete annotations represented by the global classification labels.
Such setting forms a semi-supervised learning (SSL) problem, where part of the training does not have the localisation label.
There are two typical SSL strategies, namely: pseudo-label methods~\citep{sohn2020simple,liu2021unbiased,xu2021end} and consistency learning methods~\citep{jeong2019consistency}.

\textbf{Pseudo-label methods} 
are usually explored in a student-teacher framework, where the main idea is to train the teacher network with the fully-annotated subset, followed by a training of the student model with pseudo-labels produced by the teacher for the unlabelled data. 
\citet{sohn2020simple} proposed STAC, a detection model pre-trained using labelled data and used to generate labels for unlabelled data, which is then used to fine-tune the model.
The unbiased teacher method~\citep{liu2021unbiased} generates the pseudo-labels for the student, and the student trains the teacher with EMA~\citep{tarvainen2017mean}, where the teacher and student use different augmented images, and the foreground and background imbalances are dealt with the focal loss~\citep{lin2017focal}. 
Soft teacher~\citep{xu2021end} is based on an end-to-end learning paradigm that gradually improves the pseudo-label accuracy during the training by selecting reliable predictions. 
The pseudo-labelling methods above have the advantage of being straightforward to implement for the detection problem. 
However, when fully-annotated data is scarce, the student network can overfit the incorrect pseudo-labels, which is a problem known as confirmation bias that leads to poor model performance. 

\textbf{Consistency learning methods} train a  model by minimizing the difference between the predictions produced before and after applying different types of perturbations to the unlabelled images. 
Such strategy is challenging to implement for object detection because of the difficulty to accurately match the detected regions across various locations and sizes after perturbation.
Recently proposed methods~\citep{jeong2019consistency, xu2021end} use one-to-one correspondence perturbation (such as cutout and flipping) to solve this problem. Nevertheless, such strong perturbations are not suitable for medical image detection since they could erase malignant lesions.
Also, perturbations applied to pixel values (e.g., color jitter) are 
not helpful for mammograms since they are grey-scale images. 
Furthermore, searching for the optimal combination of the perturbations and the hyper-parameters that control the data augmentation functions is computationally expensive.    
Given the drawbacks of consistency learning methods, we explore pseudo-label methods in this paper, but we introduce a mechanism to address the confirmation bias problem.

\subsection{Pre-training methods in medical image analysis}~\label{sec:liter_pretrain}

Pre-training is an important step whenever the annotated training set is too small to provide a robust model training~\citep{clancy2020deep}. 
The most successful pre-training methods are based on ImageNet pre-training~\citep{bar2015chest,carneiro2017automated}, self-supervised pre-training~\citep{vu2021improved}, or proxy-task pre-training~\citep{clancy2020deep}.

\textbf{ImageNet pre-training}~\citep{russakovsky2015imagenet} is widely used in the medical image analysis~\citep{bar2015chest,carneiro2017automated}, where it can enable faster convergence~\citep{raghu2019transfusion} or compensate for small training sets~\citep{carneiro2017automated}. 
However, ImageNet pre-training can be problematic given the fundamental differences in image characteristics between natural images and medical images.
\textbf{Self-supervised pre-training}~\citep{vu2021improved} has shown strong results, but designing an arbitrary self-supervised task that is helpful for the target task (i.e., lesion detection) is not trivial, and self-supervised pre-training tends to be a complex process that can take a significant amount of training time.
The use of \textbf{proxy-task pre-training}~\citep{clancy2020deep} is the method that usually shows the best performance, as long as the proxy task is relevant to the target task.
We follow the proxy-task pre-training given its superior performance.

\section{Proposed method}

\begin{table}[t]
    \centering
    \caption{List of Symbols.}
        \begin{tabular}{ll}
        \toprule[0.25ex]
        \multicolumn{2}{c}{Problem Definition} \\ \midrule
        $m$ & Main view of the system \\
        $a$ & Auxiliary view of the system \\
        $\mathcal{D}$ & Dataset \\
        $\mathcal{D}_s$ & Fully-annotated subset \\
        $\mathcal{D}_w$ & Weakly-annotated subset \\
        $\mathcal{Y}$ & Classification and bounding box labels of the lesions \\
        $\mathcal{C}$ & Set of classification labels \\ & (with 1 = cancer and 0 = non-cancer) \\
        $\mathcal{B}$ &         
        Domain for the top-left and bottom-right coordinates of \\ & the bounding box label \\ 
        $\mathcal{X}$ & Domain for  mammogram images\\
        \midrule\midrule
        \multicolumn{2}{c}{Pre-training on Multi-view Classification} \\ \midrule 
        $\widetilde{\mathcal{D}}_s$ & Weakly-labelled version of $\mathcal{D}_s$ \\
        $\widetilde{\mathcal{D}}_w$ & Pseudo-labelled version of $\mathcal{D}_w$ produced by GradCAM\\
        $\widetilde{\mathcal{Y}}$ & 
        Classification and bounding box pseudo labels of the  \\
        & lesions produced by MVCCL \\
        $\theta$ & MVCCL's network parameters \\
        $\gamma$ & Running mean of BatchNorm layers in MVCCL \\
        $\beta$ & Running variance of BatchNorm layers in MVCCL \\
        $\mathbf{u}$ & Feature map \\
        $\mathbf{h}$ & GradCam heatmap \\
        $\mathcal{L}$ & Set of connected components produced by $\mathbf{h}$ \\
        \midrule\midrule
        \multicolumn{2}{c}{Student-teacher SSL for Lesion Detection} \\ \midrule
        $\widehat{\mathcal{D}}_w$ & Pseudo-labelled version of $\mathcal{D}_w$ by teacher network\\
        $\widehat{\mathcal{Y}}$ & Classification and bounding box pseudo labels of the  \\
        & lesions produced by the teacher network \\
        $\theta_t$ & Model parameters for teacher network \\
        $\theta_s$ & Model parameters for student network \\
        $\lambda$ & hyper-parameter of weakly-supervised loss \\
        $\alpha$ & Smoothing factor of EMA \\
        \bottomrule[0.25ex]
        \end{tabular}
    \label{tab:symbol_list}
\end{table}

\begin{figure}[!h]
    \centering
    \includegraphics[width=1.0\linewidth]{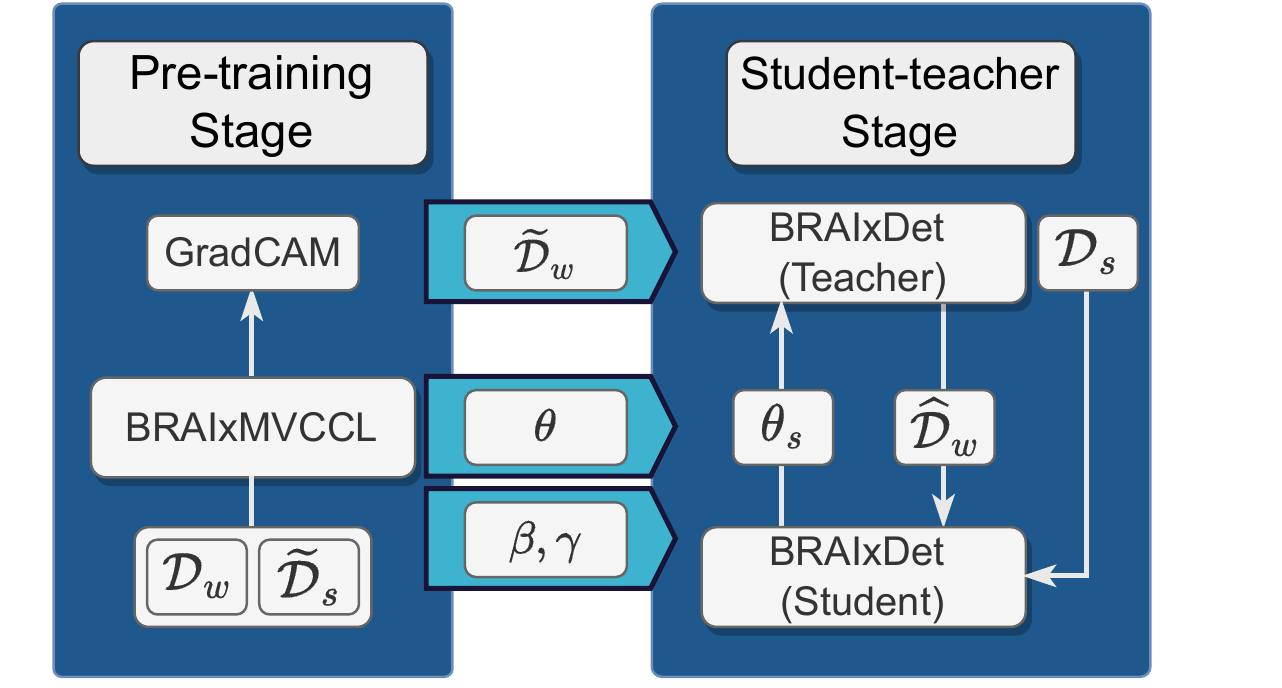}
    \caption{Overview of the proposed multi-stage training that comprises the pre-training on a weakly-supervised classification task, and a semi-supervised student-teacher learning to detect malignant lesions with incomplete annotations. During pre-training, we train the classifier BRAIxMVCCL using a weakly-labelled version of $\mathcal{D}_s$, called $\widetilde{\mathcal{D}}_s$, and  $\mathcal{D}_w$.
    After pre-training, we transfer two types of information to the next training stage: 1) the set $\widetilde{\mathcal{D}}_w$ that contains the detected lesions using the GradCAM maps from the classifier applied to the samples in $\mathcal{D}_w$; and 2) the BRAIxMVCCL's parameter $\theta$, and running mean $\gamma$ and standard deviation $\beta$ for batch norm layers.
    The semi-supervised student-teacher learning builds the teacher and student BRAIxDet detectors from BRAIxMVCCL's parameters, where the student is trained with the fully-supervised $\mathcal{D}_s$ and the pseudo-labelled $\widehat{\mathcal{D}}_w$ produced from the teacher's detections and GradCAM detections in $\widetilde{\mathcal{D}}_w$, and the teacher is trained with exponential moving average of the student's parameters.}
    \label{fig:overview}
\end{figure}

\textbf{Problem definition.}
Our method aims to train an accurate breast lesion detector from a dataset of multi-view (i.e., CC and MLO) mammograms containing incomplete annotations, i.e., a dataset 
composed of a fully-annotated subset $\mathcal{D}_s$ and a weakly-annotated subset $\mathcal{D}_w$ -- please note that all symbols used in this paper are defined in Table~\ref{tab:symbol_list}.
The fully-annotated subset is denoted by
$\mathcal{D}_s=\{\mathbf{x}_i^{m}, \mathbf{x}_i^{a}, \mathcal{Y}^m_i\}_{i=1}^{|\mathcal{D}_s|}$, where $\mathbf{x}\in\mathcal{X}\subset \mathbb{R}^{H \times W}$ represents a mammogram of height $H$ and width $W$, $\mathbf{x}_i^m$ represents the main view, $\mathbf{x}_i^a$ is the auxiliary view (with $m, a \in \{CC, MLO\}$ and $m \neq a$), $\mathcal{Y}^m_i = \{ \mathbf{c}^m_{i,j}, \mathbf{b}^m_{i,j} \}_{j=1}^{|\mathcal{Y}^m_i|}$ denotes the classification and localisation of the $|\mathcal{Y}^m_i|$ image lesions, 
with $\mathbf{c}^m_{i,j} \in \mathcal{C} = \{0, 1\}$ denoting the $j^{th}$ lesion label (with 1 = cancer and 0 = non-cancer) and $\mathbf{b}^m_{i,j} \in \mathcal{B} \in \mathbb{R}^4$ representing the top-left and bottom-right coordinates of the bounding box of the $j^{th}$ lesion on $\mathbf{x}_i^m$. The weakly-annotated subset is defined as $\mathcal{D}_w=\{\mathbf{x}_i^{m}, \mathbf{x}_i^{a}, \mathbf{c}^m_i\}_{i=1}^{|\mathcal{D}_w|}$, with $\mathbf{c}^m_i \in \mathcal{C} = \{0,1\}$.
The testing set is similarly defined as the fully-annotated subset since we aim to assess the lesion detection performance.

\textbf{System overview.}
To effectively utilise the training samples in $\mathcal{D}_s$ and $\mathcal{D}_w$, we propose a 2-stage learning process, with a weakly-supervised pre-training using all samples from $\mathcal{D}_w$ and  $\widetilde{\mathcal{D}}_s=\left\{(\mathbf{x}_i^{m}, \mathbf{x}_i^{a}, \mathbf{c}^m_i) | (\mathbf{x}_i^{m}, \mathbf{x}_i^{a}, \mathcal{Y}^m_i) \in \mathcal{D}_s \text{, and } \mathbf{c}^m_i = \max_{j\in\{1,...,|\mathcal{Y}_i^m|\}}( \mathbf{c}_{i,j}^m ) \right\}$, followed by semi-supervised student-teacher learning using  $\mathcal{D}_s$ and pseudo-labelled $\widehat{\mathcal{D}}_w$, as shown in Figure~\ref{fig:overview}.
In the \textbf{pre-training stage} described in Section~\ref{sec:pretrain}, we train the multi-view mammogram classifier BRAIxMVCCL~\citep{chen2022multi} to learn an effective feature extractor and a reasonably accurate GradCAM detector~\citep{selvaraju2017grad} to support the next stage of semi-supervised learning. 
After pre-training, we replace the classification layer of BRAIxMVCCL with a detection head, forming the BRAIxDet model based on the Faster R-CNN backbone~\citep{ren2015faster}, and we also duplicate BRAIxDet into the student and teacher models. 
In our \textbf{student-teacher semi-supervised learning (SSL)} stage explained in Section~\ref{sec:mt}, we train the student and teacher models, where the teacher uses its detector and the GradCAM detections in $\widetilde{\mathcal{D}}_w$ to generate localisation pseudo-labels, denoted by $\widehat{\mathcal{Y}}_i^m$, for the lesions in the samples $(\mathbf{x}_i^m,\mathbf{x}_i^a,\mathbf{c}_i^m) \in \mathcal{D}_w$, forming the new pseudo-labelled dataset $\widehat{\mathcal{D}}_w$ to train the student, and the student updates the teacher's model parameters based on the exponential moving average (EMA)~\citep{laine2016temporal, tarvainen2017mean} of its model parameters. 
Additionally, the student is also trained to detect lesions using the fully-annotated samples from $\mathcal{D}_s$.

During \textbf{inference}, the final lesion detection results are represented by the bounding box predictions from the \textbf{teacher} BRAIxDet model.

\subsection{Pre-training on multi-view mammogram classification}~\label{sec:pretrain}

\begin{figure*}[t]
    \centering
    \includegraphics[width=.98\textwidth]{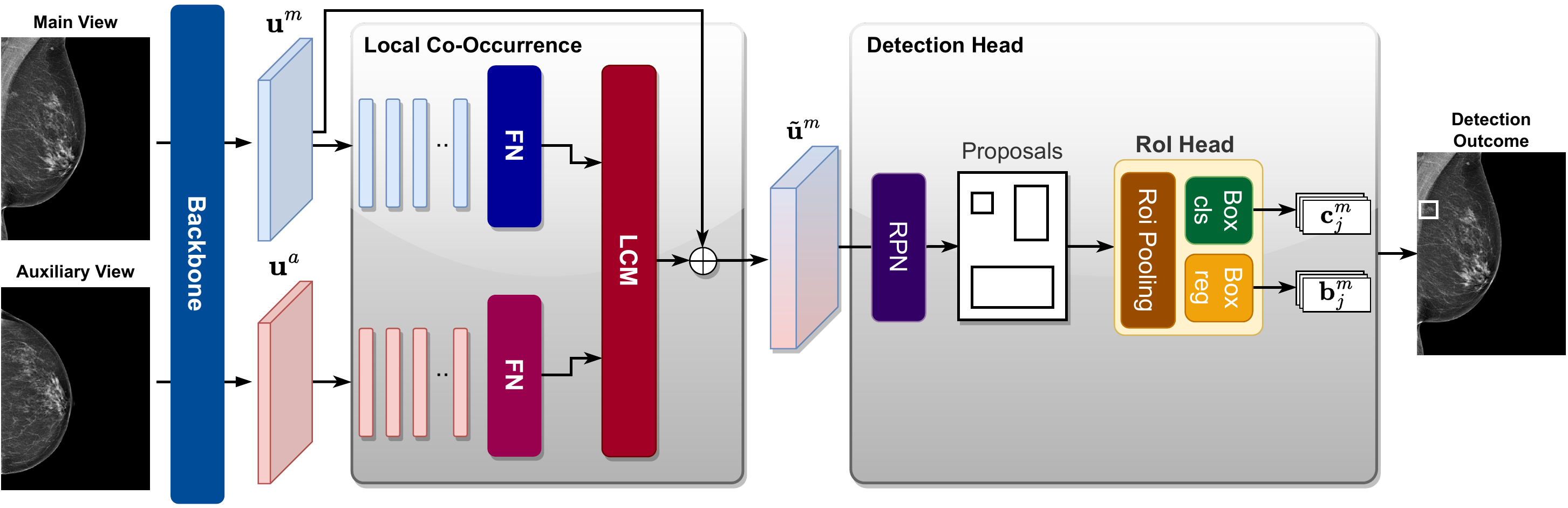}
    \caption{\textbf{BRAIxDet} takes two mammographic views (main and auxiliary) and uses a backbone model to extract the main and auxiliary features $\mathbf{u}^m$ and $\mathbf{u}^a$, where the main components are: 1) the local co-occurrence module (LCM) that models the local semantic relationships between the two views to output the cross-view feature $\tilde{\mathbf{u}}_m$; and 2) the detection module that outputs the classifications $\{\mathbf{c}^{m}_j\}_{j=1}^{|\mathcal{Y}^m|}$ and lesion bounding box predictions $\{\mathbf{b}^{m}_j\}_{j=1}^{|\mathcal{Y}^m|}$. \textbf{FN} stands for fully-connected layers, $\oplus$ indicates the concatenation operator, and \textbf{RPN} is the region proposal network. 
    }
    \label{fig:mv_backbone}
\end{figure*}

% Explain multi-view backbone
We pre-train our previously proposed BRAIxMVCCL model~\citep{chen2022multi} on the classification task before transforming it into a Faster R-CNN detector~\citep{ren2015faster} to be trained to localise cancerous lesions. 
The multi-view backbone classifier BRAIxMVCCL~\citep{chen2022multi}, parameterised by $\theta \in \Theta$ and partially illustrated in Figure~\ref{fig:mv_backbone}, is defined
by $\hat{\mathbf{c}} = f_{\theta}(\mathbf{x}^m,\mathbf{x}^a)$, which returns a classification $\hat{\mathbf{c}} \in [0,1]$ (1 = cancer and 0 = non-cancer) given the main and auxiliary views $\mathbf{x}^m,\mathbf{x}^a$. 
This model is trained with $\widetilde{\mathcal{D}}_s$ and $\mathcal{D}_w$ defined above to minimise classification mistakes and maximise the consistency between the global image features produced from each view, as follows:
\begin{equation}
\begin{split}
    \theta^* = & \argmin_{\theta \in \Theta} \\ &\sum_{(\mathbf{x}^m_i,\mathbf{x}^a_i,\mathbf{c}^m_i) \in \mathcal{D}_w \bigcup \widetilde{\mathcal{D}}_s} \ell_{bce}(f_{\theta}(\mathbf{x}^m_i,\mathbf{x}^a_i),\mathbf{c}^m_i) + \ell_{sim}(f^g_{\theta}(\mathbf{x}^m_i),f^g_{\theta}(\mathbf{x}^a_i)),
\end{split}
\end{equation}
where $\ell_{bce}(.)$ denotes the binary cross-entropy (BCE) loss, and $\ell_{sim}(.)$ denotes the consistency loss between the global features produced by $f^g_{\theta}(.)$, which is part of the model $f_{\theta}(.)$.

A particularly important structure from the BRAIxMVCCL classifier, which will be useful for the detector BRAIxDet, is the 
local co-occurrence module (LCM) that
explores the cross-view feature relationships at local regions for the main view image $\mathbf{x}^{m}$.
The LCM is denoted by $\tilde{\mathbf{u}}^m= \mathbf{u}^m \oplus f^l_{\theta}(\mathbf{u}^m,\mathbf{u}^a)$, where $\tilde{\mathbf{u}}^m \in \mathbb{R}^{\hat{H}\times\hat{W}\times D'}$ (with $\hat{H}<H$ and $\hat{W} < W$), $\oplus$ denotes the concatenation operator, and $f^l_{\theta}(.)$ is part of $f_{\theta}(.)$. 
The features used by LCM are extracted from $\mathbf{u}^m=f^b_{\theta}(\mathbf{x}^{m})$ and $\mathbf{u}^a=f^b_{\theta}(\mathbf{x}^{a})$, where
$f^b_{\theta}(.)$ is also part of $f_{\theta}(.)$ and $\mathbf{u}^m,\mathbf{u}^a \in \mathcal{U} \subset \mathbb{R}^{\hat{H}\times\hat{W}\times D}$ (with $D > D'$).

\begin{algorithm}[!t]
    \caption{Produce GradCAM pseudo-labelled dataset $\widetilde{\mathcal{D}}_w$}
    \label{alg:cam_to_box}
    \begin{algorithmic}[1]
        \State \textbf{require:} Weakly-supervised dataset $\mathcal{D}_w$, BRAIxMVCCL model $f_{\theta}(.)$, and threshold $\tau$
        \State $\widetilde{\mathcal{D}}_w=\emptyset$ \Comment{Initialise pseudo-labelled $\widetilde{\mathcal{D}}_w$}
        \For{$(\mathbf{x}_i^m,\mathbf{x}_i^a,\mathbf{c}_i) \in \mathcal{D}_w$, where $\mathbf{c}_i=1$}
        \State $\hat{\mathbf{c}}_i=f_{\theta}(\mathbf{x}_i^m,\mathbf{x}_i^a)$
            \State $\mathbf{h}_i = GradCAM(f_\theta(\mathbf{x}^m_i,\mathbf{x}^a_i), \mathbf{c}_i)$ \Comment{Heatmap prediction}
            \State $\tilde{\mathbf{h}}_i = \mathbb{I}(\mathbf{h}_i>\tau)\odot\mathbf{h}_i$
            \Comment{Binarise heatmap}
            \State $\mathcal{L}_i=CCA(\tilde{\mathbf{h}}_i)$ \Comment{Connected component analysis}
            \For{$\mathbf{l}_j \in \mathcal{L}_i$} \Comment{Remove small and large components}
                \State \textbf{if} $area(\mathbf{l}_j) < 32\times32$
                \textbf{then} $\mathcal{L}_i \leftarrow \mathcal{L}_i \setminus \mathbf{l}_j$
                \State \textbf{if}
                $area(\mathbf{l}_j) > 1024\times1024$
                \textbf{then} $\mathcal{L}_i \leftarrow \mathcal{L}_i \setminus \mathbf{l}_j$
            \EndFor
            \State $\widetilde{\mathcal{Y}}^m_i=\emptyset$ \Comment{GradCAM detections}
            \For{$\mathbf{l}_j \in \mathcal{L}_i$}
            \State $\mathbf{b}_j = BBox(\mathbf{l}_j)$ \Comment{Get bounding box from $\mathbf{l}_j$}
            \State $\widetilde{\mathcal{Y}}^m_i \leftarrow \widetilde{\mathcal{Y}}^m_i \bigcup (\hat{\mathbf{c}}_i, \mathbf{b}_j)$ \Comment{Add bounding box to $\widetilde{\mathcal{Y}}^m_i$}
            \EndFor
            \State $\widetilde{\mathcal{D}}_w \leftarrow \widetilde{\mathcal{D}}_w \bigcup (\mathbf{x}_i^m,\mathbf{x}_i^a,\widetilde{\mathcal{Y}}_i^m)$
        \EndFor
        \State \textbf{return} $\widetilde{\mathcal{D}}_w$
    \end{algorithmic}
\end{algorithm}

% GradCAM
The student-teacher SSL stage will require the detection of pseudo labels for the weakly supervised dataset $\mathcal{D}_w$.  
To avoid the confirmation bias described in Section~\ref{sec:semi-supervised_object_detection}, we combine the teacher's detection results, explained below in Section~\ref{sec:mt}, with the GradCAM~\citep{selvaraju2017grad} detections for the cancer cases produced by the BRAIxMVCCL classifier. 
Algorithm~\ref{alg:cam_to_box} displays the pseudo-code to generate GradCAM pseudo labels, where $GradCAM(f_{\theta}(\mathbf{x}_i^m,\mathbf{x}_i^a),\mathbf{c}_i)$ returns the GradCAM heatmap $\mathbf{h}_i \in [0,1]^{H \times W}$ for class $\mathbf{c}_i$ on image $\mathbf{x}_i^m$, $\mathbb{I}(\mathbf{h}_i > \tau)$ produces a binary map with the heatmap pixels larger than $\tau$ being set to 1 or 0 otherwise, $\odot$ denotes the element-wise multiplication operator, $CCA(\mathbf{h}_i)$ produces the set of connected components $\mathcal{L}_i = \{ \mathbf{l}_j \}_{j=1}^{|\mathcal{L}_i|}$ from the binarised heatmap $\tilde{\mathbf{h}}_i$ (with $\mathbf{l}_j \in \{0,1\}^{H \times W}$),  $area(\mathbf{l}_j)$ returns the area (i.e., number of pixels $\omega \in \Omega$ where $\mathbf{l}_j(\omega)=1$, where $\Omega$ is the image lattice) of the connected component $\mathbf{l}_j$, and $BBox(\mathbf{l}_j)$ returns the 4 coordinates of the bounding box $\mathbf{b}_j$ from the (left,right,top,bottom) bounds of connected component $\mathbf{l}_j$. 
This prediction process produces a set of bounding boxes paired with the confidence score $\hat{\mathbf{c}}_i$ from classifier BRAIxMVCCL~\citep{chen2022multi}, forming $\widetilde{\mathcal{Y}}^m_i = \{ (\hat{\mathbf{c}}_i,\mathbf{b}_j) \}_{j=1}^{|\widetilde{\mathcal{Y}}^m_i|}$ for each training sample.

\subsection{Student-teacher SSL for lesion detection}~\label{sec:mt}

\begin{figure*}[t]
    \centering
    \includegraphics[width=0.98\textwidth]{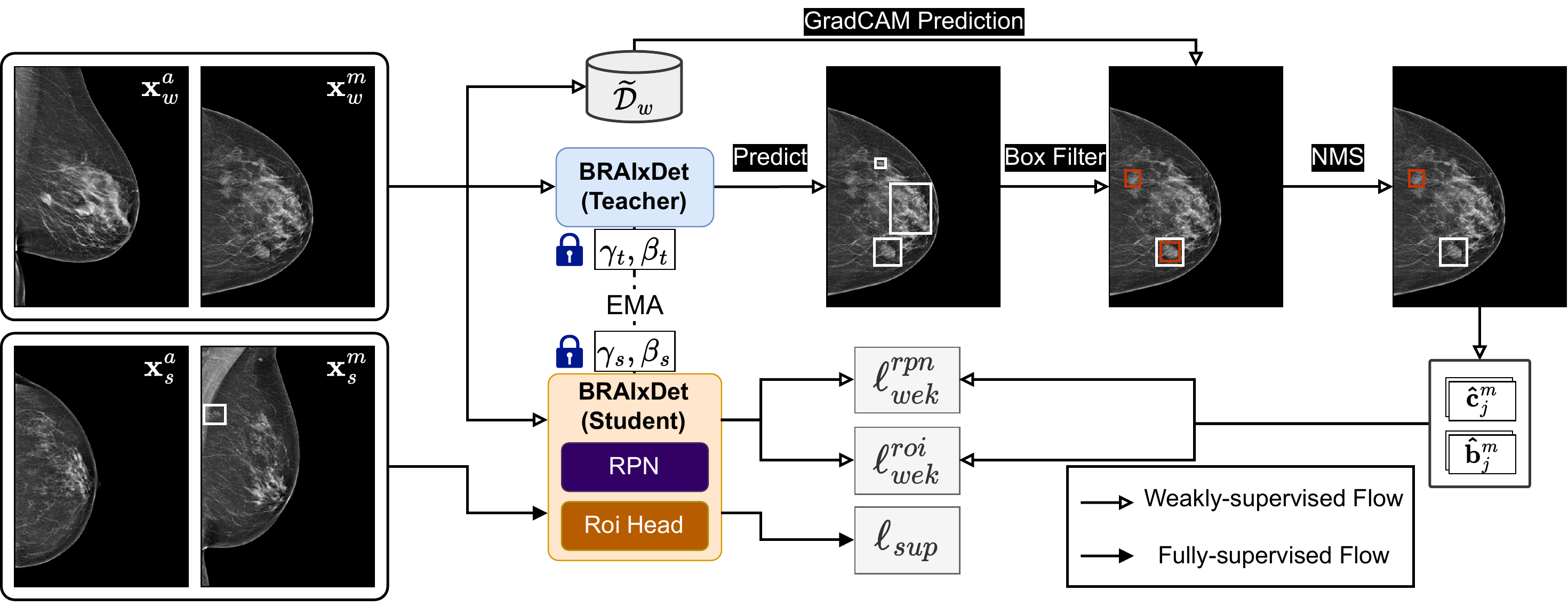}
    \caption{The student-teacher SSL of our BRAIxDet is optimised in two steps: 1) the student is trained with the fully-annotated $\mathcal{D}_s$, using loss $\ell_{sup}(.)$ in~\eqref{eq:supervised_loss_SSL}, and with the pseudo-labelled $\widehat{\mathcal{D}}_w$ produced by the teacher and GradCAM predictions in $\widetilde{\mathcal{D}}_w$ for the weakly supervised dataset $\mathcal{D}_w$; and 2) the teacher network is updated based on the EMA of the student's network parameters. 
    We introduce the following two steps to improve the quality of the pseudo labels produced by the teacher: a) a box filtering process that selects the teacher's most confident prediction; b) a non-maximum suppression (NMS) operation that rejects duplicated boxes by comparing the overlap and objectiveness score between the most confident of the teacher's predictions and the GradCAM predictions. The running mean $\gamma$ and standard deviation $\beta$ for both the batch normalisation of the teacher and student models are fixed during the entire process.}
    \label{fig:frame_work}
\end{figure*}

After pre-training, we modify the structure of the trained BRAIxMVCCL model~\citep{chen2022multi}  $f_{\theta}(\mathbf{x}_i^m,\mathbf{x}_i^a)$ by pruning the global consistency module (GCM) denoted by $f^g_{\theta}(.)$, the classification layers, and the part of the local co-occurrence module (LCM) that  produces spatial attention for the auxiliary image since these layers do not provide useful information for lesion localisation in the main image. 
This pruning process is followed by the addition of the detection head~\citep{ren2015faster}, consisting of the region proposal network (RPN) and region of interest (RoI\_head) head,
which forms the detector $\widehat{\mathcal{Y}}^m_i = f_{\theta}^d(\mathbf{x}_i^m,\mathbf{x}_i^a)$, where $\widehat{\mathcal{Y}}^m_i = \{ (\hat{\mathbf{c}}^m_{i,j},\hat{\mathbf{b}}^m_{i,j}) \}_{j=1}^{|\widehat{\mathcal{Y}}^m_i|}$, with $\hat{\mathbf{c}}^m_{i,j} \in [0,1]$ and $\hat{\mathbf{b}}^m_{i,j} \in \mathcal{B} \subset \mathbb{R}^4$. Then, we duplicate this model into a teacher $f^d_{\theta_t}(\mathbf{x}_i^m,\mathbf{x}_i^a)$ and a student $f^d_{\theta_s}(\mathbf{x}_i^m,\mathbf{x}_i^a)$ that are used in the SSL training~\citep{tarvainen2017mean}, where we train the student with the fully-labelled $\mathcal{D}_s$ and the pseudo-labelled $\widehat{\mathcal{D}}_w$, while
the teacher network is trained with the exponential moving average (EMA) of the student's parameters~\citep{tarvainen2017mean}. Figure~\ref{fig:mv_backbone} shows the BRAIxDet structure, and
Figure~\ref{fig:frame_work} summarises the student-teacher SSL training stage.

\begin{algorithm}[t]
    \caption{Non-maximum Suppression (NMS)}
    \label{alg:nms}
    \begin{algorithmic}[1]
        \State \textbf{require:} 
        % $\hat{\mathcal{Y}}_{init} = \{(b_1, s_1), ..., (b_n, s_n)\}$,
        $\mathcal{B} = \{b_1, ... ,b_n\}$,
        $\mathcal{S} = \{s_1, ..., s_n\}$,
        $\tau_{nms}$ \newline 
        % $\hat{\mathcal{Y}}_{init}:$ is the list of initially detected boxes, where each prediction contains the coordinates of the bounding box ($b_n$) and its confidence score ($s_n$).\newline
        $\mathcal{B}$ is a set of predicted bounding boxes.\newline
        $\mathcal{S}$ contains corresponding detection scores.\newline
        $\tau_{nms}:$ represent the IoU threshold.
        \State $\widehat{\mathcal{Y}} \gets \emptyset$
        \While {$\mathcal{B} \neq \emptyset$} 
        \State $i_{max} \gets \argmax_{i \in \{1,...,|\mathcal{S}|\}}    s_i \in \mathcal{S}$
        % \State $b_{max} \gets  b_{i_{max}}$
        % \State $s_{max} \gets s_{i_{max}}$
        \State $\widehat{\mathcal{Y}} \gets \widehat{\mathcal{Y}} \cup (b_{i_{max}}, s_{i_{max}})$
        \State $\mathcal{B} \gets \mathcal{B} - b_{i_{max}}$ \State $\mathcal{S} \gets \mathcal{S} - s_{i_{max}}$ 
            \For{$i \in \{1,...,|\mathcal{B}|\}$}
                \If{$IoU(b_{i_{max}}, b_{i}) \geq \tau_{nms}$} 
                    \State $\mathcal{B} \gets \mathcal{B} - b_{i}$ 
                    \State $\mathcal{S} \gets \mathcal{S} - s_{i}$
                \EndIf 
            \EndFor
        \EndWhile
        \State \textbf{return} $\widehat{\mathcal{Y}}$
    \end{algorithmic}
\end{algorithm}

For the supervised training of the student, we followed the Faster-RCNN training procedure that consists of four loss functions to train the RPN and Roi\_head modules to classify bounding boxes and regress their coordinates~\citep{ren2015faster}. The objective function is defined as
\begin{equation}
\begin{split}
    \ell_{sup} & (\mathcal{D}_s,\theta_s)  = \\ 
     & \sum_{(\mathbf{x}^{m}_i,\mathbf{x}^{a}_i,\mathcal{Y}^m_i) \in \mathcal{D}_s} \left(\ell^{rpn}_{cls}(f^d_{\theta_s}(\mathbf{x}_i^m,\mathbf{x}_i^a), \mathcal{Y}^m_i) + 
      \ell^{rpn}_{reg}(f^d_{\theta_s}(\mathbf{x}_i^m,\mathbf{x}_i^a), \mathcal{Y}^m_i)\right)  + \\
    & \sum_{(\mathbf{x}^{m}_i,\mathbf{x}^{a}_i,\mathcal{Y}^m_i) \in \mathcal{D}_s} \left(\ell^{roi}_{cls}(f^d_{\theta_s}(\mathbf{x}_i^m,\mathbf{x}_i^a), \mathcal{Y}^m_i) + 
      \ell^{roi}_{reg}(f^d_{\theta_s}(\mathbf{x}_i^m,\mathbf{x}_i^a), \mathcal{Y}^m_i)\right),
\end{split}
\label{eq:supervised_loss_SSL}
\end{equation}
where $\ell^{rpn}_{cls}(.)$, $\ell^{rpn}_{reg}(.)$ denote the RPN classification and regression losses with respect to the anchors, and $\ell^{roi}_{cls}(.)$, $\ell^{roi}_{reg}(.)$ represent the RoI\_head classification and regression losses with respect to the region proposal features~\citep{ren2015faster}. 

These pseudo-labels to train the student are provided by the teacher, which form the pseudo-labelled dataset $\widehat{\mathcal{D}}_w = \left\{ (\mathbf{x}_i^m,\mathbf{x}_i^a,\widehat{\mathcal{Y}}_i^m)| (\mathbf{x}_i^m,\mathbf{x}_i^a,\mathbf{c}_i^m) \in \mathcal{D}_w \text{, and } \widehat{\mathcal{Y}}_i^m = f^d_{\theta_t}(\mathbf{x}_i^m,\mathbf{x}_i^a) \right\}$.
However, at the beginning of the training (first two training epochs), the pseudo-labels in $\widehat{\mathcal{D}}_w$ are unreliable, which can cause the confirmation bias issue mentioned in Section~\ref{sec:semi-supervised_object_detection}, so we also use the GradCAM predictions in $\widetilde{\mathcal{D}}_w$ from Algorithm~\ref{alg:cam_to_box}.
More specifically, starting from $(\mathbf{x}_i^m,\mathbf{x}_i^a,\widehat{\mathcal{Y}}_i^m) \in \widehat{\mathcal{D}}_w$, we first select the most confident prediction from $(\hat{\mathbf{c}}^m_{i,j^*},\hat{\mathbf{b}}_{i,j^*}^m) \in \widehat{\mathcal{Y}}_i^m$, with $j^* = \argmax_{j\in\{1,...,|\widehat{\mathcal{Y}}_i^m|\}}\hat{\mathbf{c}}_{i,j}$, and add it to the set of GradCAM detections in $\widetilde{\mathcal{Y}}_i^m$.
Then, for each training sample $(\mathbf{x}_i^m,\mathbf{x}_i^a,\widehat{\mathcal{Y}}_i^m) \in \widehat{\mathcal{D}}_w$ we run non-max suppression (NMS, as shown in Algorithm~\ref{alg:nms}) on the teacher and GradCAM detections to form $\widehat{\mathcal{Y}}_i^m \leftarrow NMS(\widetilde{\mathcal{Y}}_i^m \bigcup (\hat{\mathbf{c}}^m_{i,j^*},\hat{\mathbf{b}}_{i,j^*}^m))$.
After these initial training stages, the pseudo-label is solely based on the teacher's prediction from $\widehat{\mathcal{D}}_w$.
The objective function to train the student using $\widehat{\mathcal{D}}_w$ is defined as
\begin{equation}
    \begin{split}
    \ell_{wek} & (\widehat{\mathcal{D}}_w,\theta_s) = \\
   & \sum_{(\mathbf{x}^{m}_i,\mathbf{x}^{a}_i,\widehat{\mathcal{Y}}^m_i) \in \widehat{\mathcal{D}}_w} \left(\ell^{rpn}_{cls}(f^d_{\theta_s}(\mathbf{x}_i^m,\mathbf{x}_i^a), \widehat{\mathcal{Y}}^m_i) + 
      \ell^{rpn}_{reg}(f^d_{\theta_s}(\mathbf{x}_i^m,\mathbf{x}_i^a), \widehat{\mathcal{Y}}^m_i)\right)  + \\
    & \sum_{(\mathbf{x}^{m}_i,\mathbf{x}^{a}_i,\widehat{\mathcal{Y}}^m_i) \in \widehat{\mathcal{D}}_w} \left(\ell^{roi}_{cls}(f^d_{\theta_s}(\mathbf{x}_i^m,\mathbf{x}_i^a), \widehat{\mathcal{Y}}^m_i) + 
      \ell^{roi}_{reg}(f^d_{\theta_s}(\mathbf{x}_i^m,\mathbf{x}_i^a), \widehat{\mathcal{Y}}^m_i)\right),
    \end{split}
\end{equation}
where the losses used here are the same as the ones  in~\eqref{eq:supervised_loss_SSL}.

The overall loss function to train the student is 
\begin{equation}
    \ell_{stu}(\mathcal{D}_s, \widehat{\mathcal{D}}_w,\theta_s) = \ell_{sup}(\mathcal{D}_s,\theta_s) + \lambda \ell_{wek}(\widehat{\mathcal{D}}_w,\theta_s),
    \label{eq:overall_student_loss}
\end{equation}
where $\lambda$ is the hyper-parameter that controls the contribution of the weakly-supervised loss.
The teacher's model parameters are updated with EMA~\citep{tarvainen2017mean}:
\begin{equation}
    \theta'_t = \alpha \theta_{t} + (1-\alpha) \theta_s,
    \label{eq:teacher_ema_loss}
\end{equation}
where $\alpha\in(0, 1)$ is a hyper-parameter that represents the smoothing factor, where a high value of $\alpha$ provides a slow updating process.

\subsection{Batch Normalisation for EMA}
In the original mean-teacher training~\citep{tarvainen2017mean}, both student and teacher use the standard batch normalization (BN). 
For the student network, the model parameter $\theta_s$ is aligned with the BN parameters $\gamma$ and standard deviation $\beta$ since these parameters are optimized based on the batch-wise statistics. 
However, this relationship does not hold for the teacher's network since the teacher's model parameter $\theta_t$ is updated by EMA, but the BN statistics are not updated. 
\citet{cai2021exponential} claimed that this issue can be solved by applying EMA on BN statistics to avoid the misalignment in the teacher's parameter space. 
We propose a simpler approach to address this issue, which is to simply freeze the BN layers for both student and teacher models since the BN statistics is already well estimated from the entire dataset $\widetilde{\mathcal{D}}_s \bigcup \mathcal{D}_w$ during pre-training. 
We show in the experiments that our proposal above addresses well the mismatch between the student's and the teacher's parameters and the dependency on the student's training samples, providing better detection generalisation than presented by \citet{cai2021exponential}.

\section{Experiments}

In this section, we first introduce the two datasets used in the experiments, and then we explain the experimental setting containing variable proportions of fully and weakly annotated samples in the training set. Next, we discuss the implementation details of our method and competing approaches. We conclude the section with a visualization of the detection results and ablation studies.

\subsection{Datasets}

We validate our proposed BRAIxDet method on two datasets that contain incomplete malignant breast lesion annotations, namely: our own Annotated Digital Mammograms and Associated Non-Image data (ADMANI), and the public Curated Breast Imaging Subset of the Digital Database for Screening Mammography
(CBIS-DDSM)~\citep{lee2017curated}.

%INBreast dataset~\citep{moreira2012inbreast}.

The \textbf{ADMANI Dataset}
was collected from several breast screening clinics from the State of Victoria in Australia, between 2013 and 2019, and contains pre-defined training and testing sets. 
Each exam on ADMANI has two mammographic views (CC and MLO) per breast produced by one of the following manufactures: Siemens\textsuperscript{TM}, Hologic\textsuperscript{TM}, Fujifilm Corporation\textsuperscript{TM}, Philips\textsuperscript{TM} Digital Mammography Sweden AB, Konica Minolta\textsuperscript{TM}, GE\textsuperscript{TM} Medical Systems, Philips Medical Systems\textsuperscript{TM}, and Agfa\textsuperscript{TM}. 
The training set contains 771,542 exams with 15,994 cancer cases (containing malignant findings) and 3,070,174 non-cancer cases (with 44,040 benign cases and 3,026,134 cases with no findings). This training set is split 90/10 for training/validation in a patient-wise manner.
The training set has 7,532 weakly annotated cancer cases and 6,892 fully annotated cancer cases, while the validation set has 759 fully annotated cancer cases and no weakly annotated cases since they are not useful for model selection.
The testing set contains 83,990 exams with 1,262 cancer cases (containing malignant findings) and 334,698 non-cancer cases (with 3,880 benign cases and 330,818 no findings). 
Given that we are testing lesion detection, we remove all non-cancer cases, and all weakly annotated cancer cases, leaving us with 900 fully annotated cancer cases.

% ADMANI --- 1/16, 1/8, 1/4, 1/2, 100%+EXTRA
\begin{table*}[t]
    \centering
    \caption{mAP and Recall @ 0.5 results on the testing images of ADMANI~\citep{frazer2022admani} using the \underline{partially labelled protocol} based on splitting the fully annotated subset with the ratios 1/16, 1/8, 1/4, 1/2, and 3/4 (we show the number of fully annotated images inside the brackets). 
    We highlight the best result in each column. The last row shows the $p$-values from the one-tailed t-test to compare our BRAIxDet against the second best method (Soft Teacher~\citep{xu2021end}).}
    \label{tab:admani_partial_result}
    \color{r1}
    \resizebox{1\textwidth}{!}{  
    \begin{tabular}{l||c|c|c|c|c||c|c|c|c|c}
    \toprule																			
    \multirow{2}{*}{Method}	&	\multicolumn{5}{c||}{mAP}									&	\multicolumn{5}{c}{Recall @ 0.5}									\\	\cmidrule{2-11}
    	&	1/16 (430)	&	1/8 (861)	&	1/4 (1723)	&	1/2 (3446)	&	3/4 (5169)	&	1/16 (430)	&	1/8 (861)	&	1/4 (1723)	&	1/2 (3446)	&	3/4 (5169)	\\	\midrule
    Faster RCNN~\citep{ren2015faster}	&	61.88 ± 0.64	&	70.63 ± 0.71	&	75.49 ± 0.74	&	80.87 ± 0.81	&	82.01 ± 0.67	&	62.33 ± 0.80	&	72.15 ± 0.52	&	76.77 ± 0.58	&	81.41 ± 0.76	&	82.27 ± 0.78	\\	\midrule
    CVR-RCNN~\citep{ma2021cross}	&	62.99 ± 0.83	&	71.13 ± 0.80	&	76.32 ± 0.69	&	81.94 ± 0.86	&	83.09 ± 0.93	&	63.29 ± 0.67	&	73.17 ± 0.66	&	77.48 ± 0.57	&	82.55 ± 0.64	&	83.25 ± 0.70	\\	\midrule
    MommiNet-V2~\citep{yang2021momminetv2}	&	63.64 ± 0.75	&	72.51 ± 0.93	&	77.21 ± 0.76	&	82.52 ± 0.79	&	84.07 ± 0.81	&	63.72 ± 0.93	&	74.25 ± 0.59	&	78.05 ± 0.89	&	83.48 ± 0.75	&	83.85 ± 0.67	\\	\midrule
    STAC~\citep{sohn2020simple}	&	73.89 ± 0.80	&	79.18 ± 0.95	&	81.98 ± 0.89	&	84.32 ± 0.71	&	85.21 ± 0.76	&	74.70 ± 0.81	&	82.66 ± 0.77	&	87.13 ± 0.64	&	88.16 ± 0.52	&	89.42 ± 0.58	\\	\midrule
    MT~\citep{tarvainen2017mean}	&	78.21 ± 0.81	&	78.96 ± 0.99	&	85.78 ± 0.82	&	86.52 ± 0.75	&	87.82 ± 0.69	&	81.66 ± 0.92	&	85.50 ± 0.60	&	90.67 ± 0.76	&	90.30 ± 0.62	&	91.12 ± 0.68	\\	\midrule
    Unbiased Teacher~\citep{liu2021unbiased}	&	63.57 ± 0.67	&	75.87 ± 0.72	&	81.04 ± 0.61	&	82.17 ± 0.83	&	83.82 ± 0.68	&	64.31 ± 0.71	&	79.41 ± 0.59	&	84.38 ± 0.61	&	85.05 ± 0.74	&	86.23 ± 0.72	\\	\midrule
    Soft Teacher~\citep{xu2021end}	&	81.84 ± 0.91	&	83.46 ± 0.83	&	85.96 ± 0.75	&	86.75 ± 0.89	&	87.74 ± 0.64	&	85.92 ± 0.59	&	89.38 ± 0.67	&	89.98 ± 0.61	&	90.69 ± 0.79	&	91.77 ± 0.59	\\	\midrule
    BRAIxDet	&	\textbf{88.34 ± 0.61}	&	\textbf{89.10 ± 0.62}	&	\textbf{90.57 ± 0.79}	&	\textbf{91.84 ± 0.75}	&	\textbf{92.33 ± 0.62}	&	\textbf{91.21 ± 0.46}	&	\textbf{92.73 ± 0.74}	&	\textbf{93.58 ± 0.56}	&	\textbf{94.08 ± 0.66}	&	\textbf{94.88 ± 0.57}	\\	\midrule
     BRAIxDet (p-value)	&	$<$ 1e-14	&	$<$ 1e-14	&	$<$ 1e-14	&	$<$ 1e-14	&	$<$ 1e-14	&	$<$ 1e-14	&	$<$ 1e-14	&	$<$ 1e-14	&	$<$ 1e-14	&	$<$ 1e-14	\\	\bottomrule
    \end{tabular}
    }
\end{table*}
\begin{table}[t]
    \centering
    \caption{mAP and Recall @ 0.5 results on the testing images of ADMANI~\citep{frazer2022admani} using the \underline{fully labelled protocol} based on all fully annotated data and extra weakly-labelled data (we show the number of extra weakly labelled images inside the brackets). We highlight the best result in each column.}
    \label{tab:admani_full_result}
    \resizebox{1\linewidth}{!}{  
    \begin{tabular}{l||C{0.1\textwidth}|C{0.1\textwidth}}
    \toprule						
    \multirow{2}{*}{Method}	&	\multicolumn{2}{c}{100\%+Extra (6298)}	\\	\cmidrule{2-3}
    	&	mAP	&	Recall @ 0.5	\\	\midrule
    Faster RCNN~\citep{ren2015faster}	&	84.52 ± 0.65	&	85.09 ± 0.69	\\	\midrule
    CVR-RCNN~\citep{ma2021cross}	&	85.52 ± 0.60	&	86.03 ± 0.84	\\	\midrule
    MommiNet-V2~\citep{yang2021momminetv2}	&	85.91 ± 0.89	&	86.87 ± 0.95	\\	\midrule
    STAC~\citep{sohn2020simple}	&	88.82 ± 0.78	&	92.49 ± 0.58	\\	\midrule
    MT~\citep{tarvainen2017mean}	&	89.11 ± 1.01	&	92.22 ± 0.62	\\	\midrule
    Unbiased Teacher~\citep{liu2021unbiased}	&	87.17 ± 0.61	&	92.48 ± 0.95	\\	\midrule
    Soft Teacher~\citep{xu2021end}	&	90.29 ± 0.88	&	92.71 ± 0.91	\\	\midrule
    BRAIxDet	&	\textbf{92.78 ± 0.78}	&	\textbf{95.18 ± 0.68}	\\	\bottomrule						
    \end{tabular}
    }
\end{table}

% DDSM
% DDSM
\begin{table*}[t]
% \color{r1}
    \centering
    \caption{mAP and Recall @ 0.5 results on the testing images of CBIS-DDSM~\citep{web:ddsm} using the \underline{partially labelled protocol} based on splitting the fully annotated subset with the ratios 1/16, 1/8, 1/4, 1/2, 3/4 (we show the number of fully annotated images inside the brackets). We highlight the best result in each column. The last row shows the $p$-values from the one-tailed t-test to compare our BRAIxDet against the second best method (Soft Teacher~\citep{xu2021end}).}
    \label{tab:ddsm_partial_result}
\color{r1}
    \resizebox{1.0\textwidth}{!}{  
    \begin{tabular}{l||c|c|c|c|c||c|c|c|c|c}
\toprule																							
\multirow{2}{*}{Method}	&	\multicolumn{5}{c||}{mAP}									&	\multicolumn{5}{c}{Recall @ 0.5}									\\	\cmidrule{2-11}	
	&	1/16 (65)	&	1/8 (130)	&	1/4 (260)	&	1/2 (520)	&	3/4 (780)	&	1/16 (65)	&	1/8 (130)	&	1/4 (260)	&	1/2 (520)	&	3/4 (780)	\\	\midrule	
Faster RCNN~\citep{ren2015faster}	&	24.68 ± 0.52	&	42.35 ± 0.75	&	49.55 ± 0.47	&	54.33 ± 0.72	&	56.01 ± 0.65	&	36.78 ± 0.57	&	54.35 ± 0.66	&	61.55 ± 0.59 	&	66.64 ± 0.61	&	70.89 ± 0.74	\\	\midrule	
CVR-RCNN~\citep{ma2021cross}	&	25.86 ± 0.63	&	42.94 ± 0.88	&	50.05 ± 0.81	&	55.12 ± 0.63	&	61.92 ± 0.82	&	37.12 ± 0.89	&	55.56 ± 0.36	&	62.32 ± 0.67	&	70.39 ± 0.58	&	72.52 ± 0.67	\\	\midrule	
MommiNet-V2~\citep{yang2021momminetv2}	&	26.22 ± 0.79	&	43.75 ± 0.84	&	50.82 ± 0.56	&	59.68 ± 0.71	&	65.88 ± 0.66	&	37.94 ± 0.76	&	56.22 ± 0.62	&	63.32 ± 0.87	&	72.73 ± 0.78	&	75.62 ± 0.65	\\	\midrule	
MT~\citep{tarvainen2017mean}	&	23.31 ± 0.51	&	33.77 ± 0.45	&	56.61 ± 0.63	&	64.59 ± 0.78	&	66.22 ± 0.82	&	44.02 ± 0.94	&	45.32 ± 0.81	&	64.48 ± 0.73	&	74.05 ± 0.82	&	75.94 ± 0.69	\\	\midrule	
STAC~\citep{sohn2020simple}	&	29.16 ± 0.62	&	45.91 ± 0.73	&	56.78 ± 0.67	&	65.54 ± 0.65	&	66.47 ± 0.77	&	45.05 ± 0.74	&	61.65 ± 0.92	&	71.64 ± 0.65	&	77.58 ± 0.89	&	78.15 ± 0.82	\\	\midrule	
Unbiased Teacher~\citep{liu2021unbiased}	&	36.19 ± 0.66	&	50.38 ± 0.57	&	58.83 ± 0.81	&	65.05 ± 0.51	&	67.52 ± 0.58	&	53.78 ± 0.58	&	62.39 ± 0.61	&	69.18 ± 0.51	&	76.18 ± 0.62	&	77.47 ± 0.55	\\	\midrule	
Soft Teacher~\citep{xu2021end}	&	45.59 ± 0.70	&	48.96 ± 0.85	&	62.03 ± 0.91	&	67.61 ± 0.80	&	68.12 ± 0.49	&	57.91 ± 0.73	&	65.98 ± 0.66	&	71.27 ± 0.58	&	79.20 ± 0.71	&	81.12 ± 0.82	\\	\midrule	
BRAIxDet	&	\textbf{53.99 ± 0.83}	&	\textbf{65.29 ± 0.61}	&	\textbf{69.32 ± 0.78}	&	\textbf{72.32 ± 0.66}	&	\textbf{73.38 ± 0.72}	&	\textbf{65.84 ± 0.91}	&	\textbf{72.50 ± 0.44}	&	\textbf{76.45 ± 0.68}	&	\textbf{81.82 ± 0.79}	&	\textbf{83.16 ± 0.57}	\\	\midrule	
BRAIxDet (p-value)	&	$<$ 1e-14	&	$<$ 1e-14	&	$<$ 1e-14	&	$<$ 1e-14	&	$<$ 1e-14	&	$<$ 1e-14	&	$<$ 1e-14	&	$<$ 1e-14	&	0.0005	&	0.0002	\\	\bottomrule	
    \end{tabular}
    }
\end{table*}

% DDSM
\begin{table}[t]
    \centering
    \caption{mAP and Recall @ 0.5 results on the testing images of ADMANI~\citep{frazer2022admani} and CBIS-DDSM~\citep{web:ddsm} using only the \underline{supervised data}.}
    \label{tab:supervised}
\color{r1}
    \resizebox{1\linewidth}{!}{  
    \begin{tabular}{l||c|c||c|c}
\toprule													
~	&	\multicolumn{2}{c||}{ADMANI}					&	\multicolumn{2}{c}{DDSM}				\\	\midrule
Method	&	mAP		&	Recall @ 0.5		&	mAP		&	Recall @ 0.5	\\	\midrule
Faster RCNN~\citep{ren2015faster}	&	84.52 ± 0.65		&	85.09 ± 0.69		&	64.29 ± 0.77		&	76.86 ± 0.69	\\	\midrule
CVR-RCNN~\citep{ma2021cross}	&	85.52 ± 0.60		&	86.03 ± 0.84		&	69.75 ± 0.78		&	82.01 ± 0.57	\\	\midrule
MommiNet-V2~\citep{yang2021momminetv2}	&	85.91 ± 0.89		&	86.87 ± 0.95		&	72.65 ± 0.81		&	83.40 ± 0.74	\\	\midrule
BRAIxDet	&	\textbf{88.68 ± 0.57}		&	\textbf{91.48 ± 0.74}		&	\textbf{74.79 ± 0.62}		&	\textbf{84.88 ± 0.71}	\\	\midrule
BRAIxDet (p-value)	&	0.0009		&	2e-10		&	0.0003		&	0.0186	\\	\bottomrule
    \end{tabular}
    }
\end{table}

\begin{table}[!t]
    \centering
    \caption{Influence of each stage of our method using mAP and Recall @ 0.5 results on ADMANI dataset~\citep{frazer2022admani} under fully-labelled protocol.
    \textbf{$1^{st}$ row:} Original GradCAM result.
    \textbf{$2^{nd}$ row:} Faster RCNN~\citep{ren2015faster} with its EfficientNet-b0 backbone pre-trained with fully- and weakly-annotated subsets and trained
    using only the fully-annotated subset. 
    \textbf{$3^{rd}$ row:} pre-train the multi-view BRAIxMVCCL classifier~\citep{chen2022multi}, and keep the LCM module for the second training stage that uses the fully-annotated training subset. 
    \textbf{$4^{th}$ row:} train Faster RCNN with the student-teacher SSL using a backbone without LCM. 
    \textbf{$5^{th}$ row:} integrate Faster RCNN, LCM and student-teacher SSL. \textbf{$6^{th}$ row:} BRAIxDet results using the BRAIxMVCCL's GradCAM predictions.}
    \label{tab:ablation}
\color{r1}
    \resizebox{1\linewidth}{!}{
    \begin{tabular}{cccc||c||c}
    \toprule												
    Faster RCNN	&	MV	&	Studen-teacher	&	CAM	&	mAP	&	Recall @ 0.5	\\	\midrule
    ~	&		&		&	\checkmark	&	34.23 ± 0.72	&	38.18 ± 0.94 	\\
    \checkmark	&		&		&		&	84.52 ± 0.65	&	85.09 ± 0.69	\\	
    \checkmark	&	\checkmark	&		&		&	90.05 ± 0.89	&	93.54 ± 0.78	\\	
    \checkmark	&		&	\checkmark	&		&	89.38 ± 0.75	&	91.51 ± 0.85	\\	
    \checkmark	&	\checkmark	&	\checkmark	&		&	90.51 ± 0.64	&	93.22 ± 0.63	\\	
    \checkmark	&	\checkmark	&	\checkmark	&	\checkmark	&	\textbf{92.78 ± 0.78}	&	\textbf{95.18 ± 0.68}	\\	\bottomrule
    \end{tabular}
    }
\end{table}

The \textbf{publicly available CBIS-DDSM dataset}~\citep{lee2017curated} contains images from 1,566 participants, where 1,457 cases have malignant findings (with 1,696 mass cases and 1,872 calcification cases). 
The dataset has a pre-defined training and testing split, with 2,864 images (with 1,181 malignant cases and 1,683 benign cases) and 704 (with 276 malignant cases and 428 benign cases) images respectively. 
Each exam on CBIS-DDSM also has two mammographic views (CC and MLO) per breast.

\subsection{Experimental settings}
% Semi-setup
To systematically test the robustness of our method to different rates of incomplete malignant breast lesion annotations, we propose an experimental setting that contains different proportions of fully- and weakly-annotated samples.
In the \textbf{partially labelled protocol}, we follow typical semi-supervised settings~\citep{sohn2020simple,liu2021unbiased,xu2021end,liu2022perturbed}, where we sub-sample the fully-annotated subset using the ratio $1/n$, where the remaining $1-1/n$ of the subset becomes weakly-annotated. More specifically, on ADMANI, we split the fully-annotated subset with the ratios $3/4$, $1/2$, $1/4$, $1/8$, $1/16$. We adopt the same experimental setting for CBIS-DDSM~\citep{lee2017curated}.
In the \textbf{fully labelled protocol}, we use all ADMANI training samples that have the lesion localisation annotation as the fully-annotated subset and 
all remaining weakly-labelled samples as the weakly-annotated subset. 
Unlike the synthetic partially labelled protocol setting, the fully labelled protocol setup is a real-world challenge since it uses all data available from ADMANI, allowing methods to leverage all the available samples to improve detector performance on a large-scale mammogram dataset.

%\textbf{Evaluation metrics}
All methods are assessed using the standard \textbf{mean average precision (mAP)} and \textbf{free-response receiver operating characteristic (FROC)}.
The mAP measures the average precision of true positive (TP) detections of malignant lesions, where a TP detection is defined as producing at least a $0.2$ intersection over union (IoU) with respect to the bounding box annotation~\citep{liu2020cross,yang2020momminet,yang2021momminetv2}. We follow previous works~\citep{dhungel2017deep,agarwal2020deep, liu2019unilateral,liu2020cross,yang2021momminetv2} when selecting a threshold of 0.2 for testing, due to the following reasons: 1) large overlaps between annotations and detections that do not happen as often as in natural images since mammograms represent a compressed and thin 3D space compared to natural images~\citep{ribli2018detecting}, resulting in lesions that occupy a rather small region of the image; and 
2) inconsistent annotation quality where ground-truth labels can be much larger/smaller than its actual size, which may cause inaccurate evaluation.
On the other hand, FROC measures the recall at different false positive detections per image (FPPI). In this paper, we measure the recall at 0.5 FPPI (Recall @ 0.5), which means the recall at which the detector produces one false positive every two images. Such Recall@0.5 is a common measure to assess lesion detection from mammograms~\citep{ma2021cross,yang2020momminet, yang2021momminetv2}.

\begin{figure*}[ht]
    \color{r2}
    \subfigure[Ablation study of $\alpha$]{\includegraphics[width=0.32\linewidth]{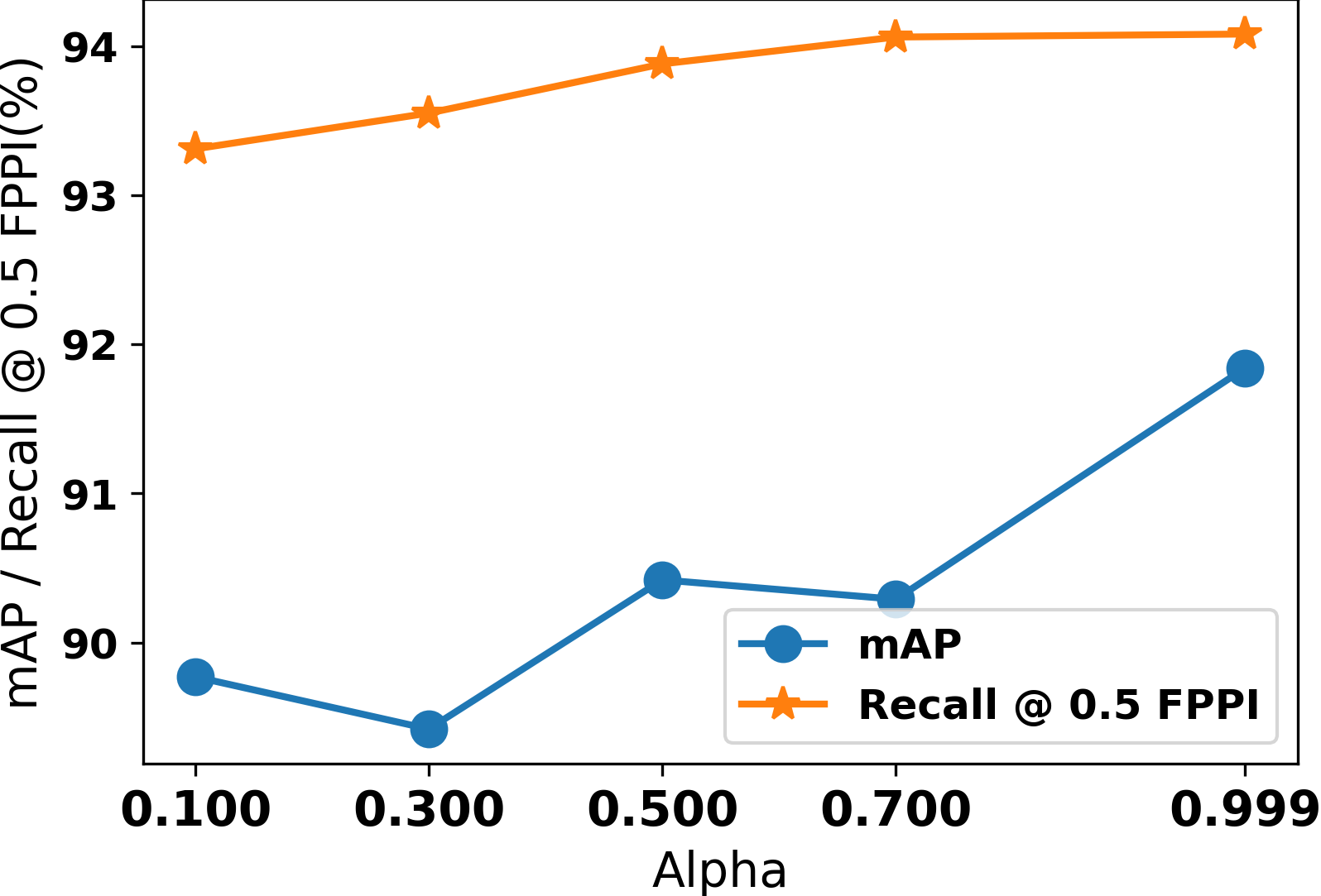}\label{fig:ablation_alpha}}
    \subfigure[Ablation study of $\tau$]{\includegraphics[width=0.32\linewidth]{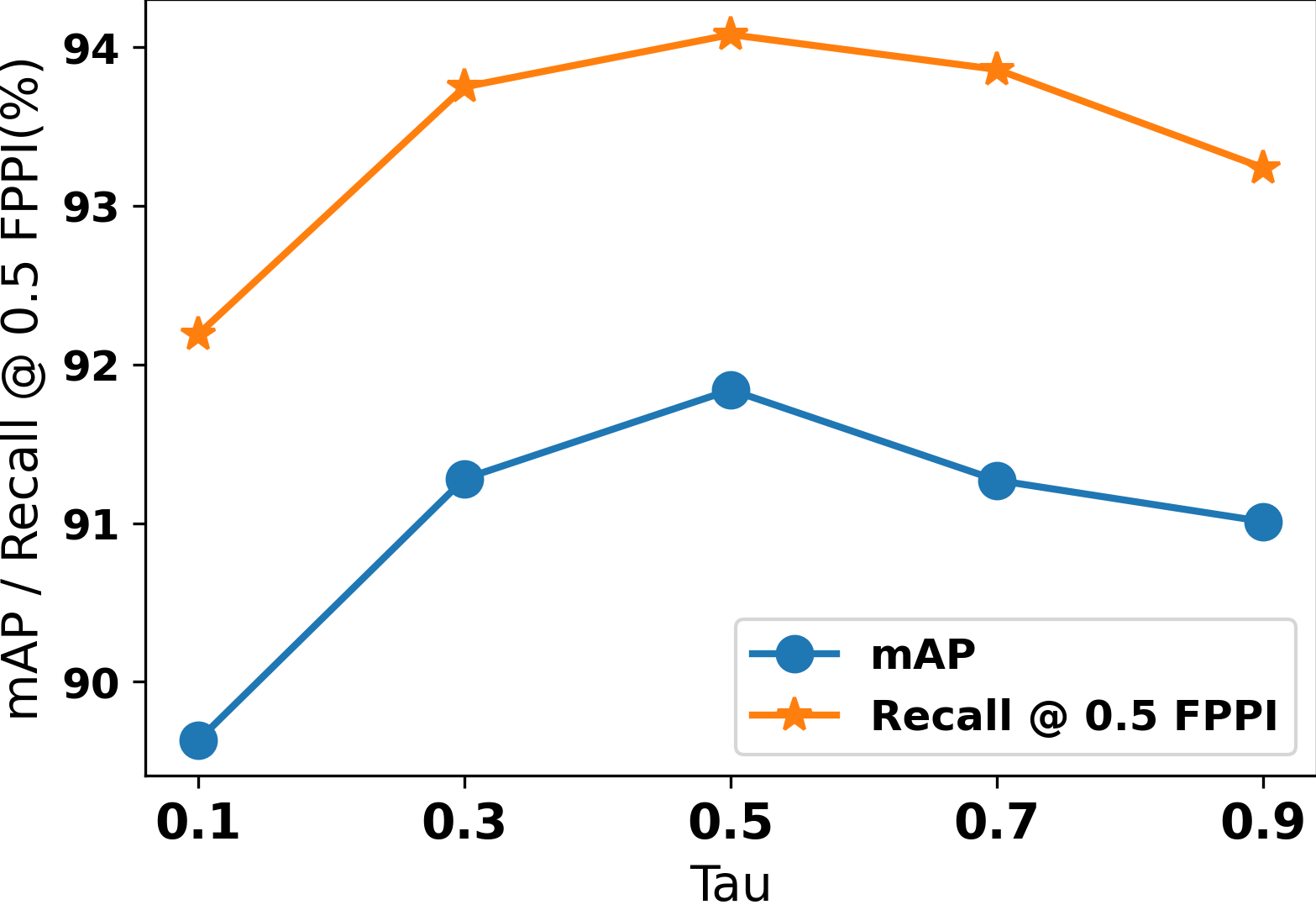}\label{fig:ablation_tau}}
    \subfigure[Ablation study of $\lambda$]{\includegraphics[width=0.32\linewidth]{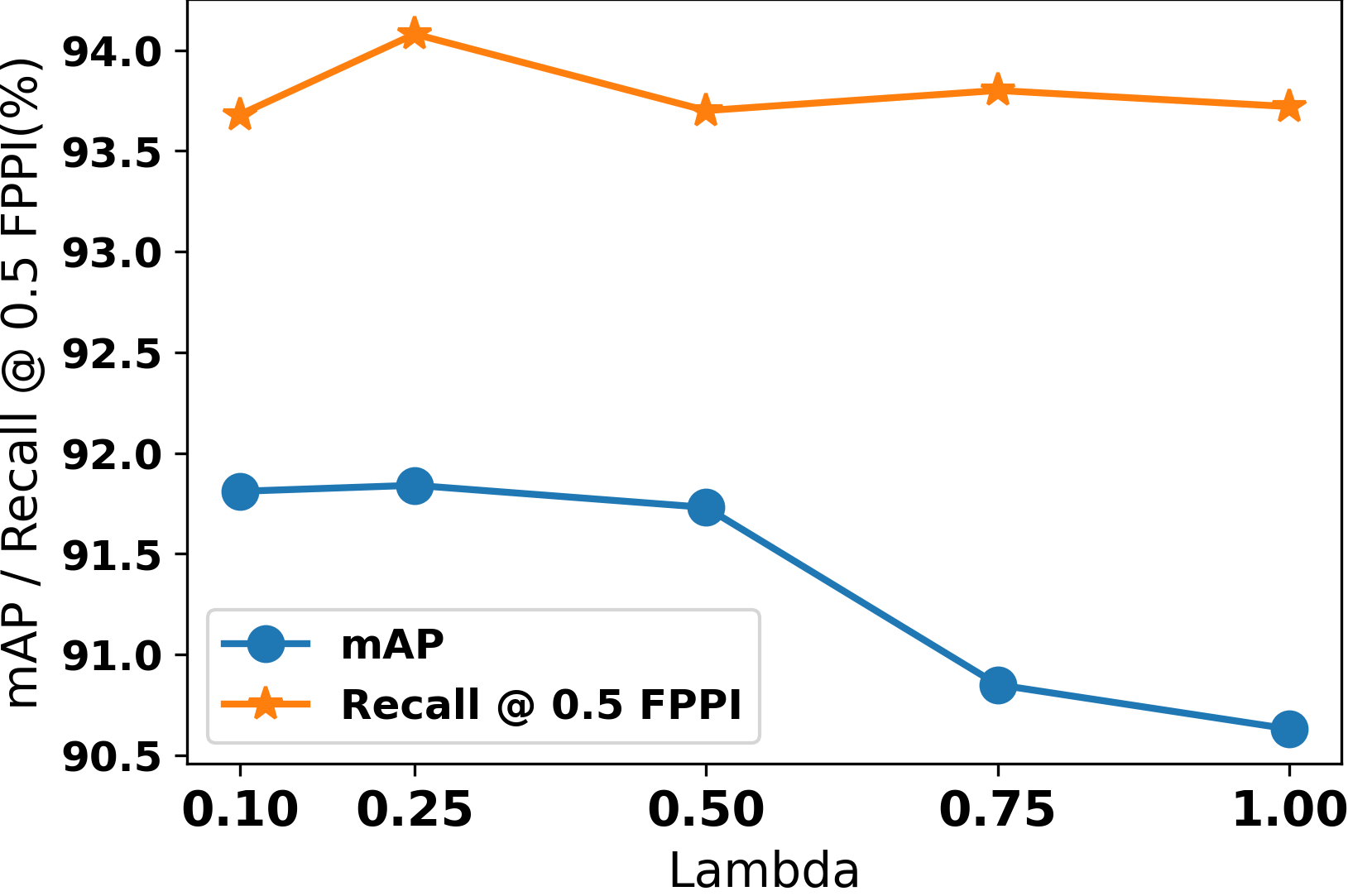}\label{fig:ablation_lambda}}
    \caption{Ablation study of BRAIxDet's hyper-parameters using mAP and Recall @ 0.5 results on ADMANI dataset~\citep{frazer2022admani} under 50\% partially labelled protocol. The hyper-parameters include: \blueb{(a)} the smoothing factor $\alpha$ in EMA process; \blueb{(b)} the threshold $\tau$ used to binarise the heatmap (Alg. 1); and \blueb{(c)} the weighting factor $\lambda$ that controls the contribution of the weakly supervised loss.}
    \label{fig:ablation_hyperp}
\end{figure*}

\begin{figure}[t]
    \centering
    \includegraphics[width=\linewidth]{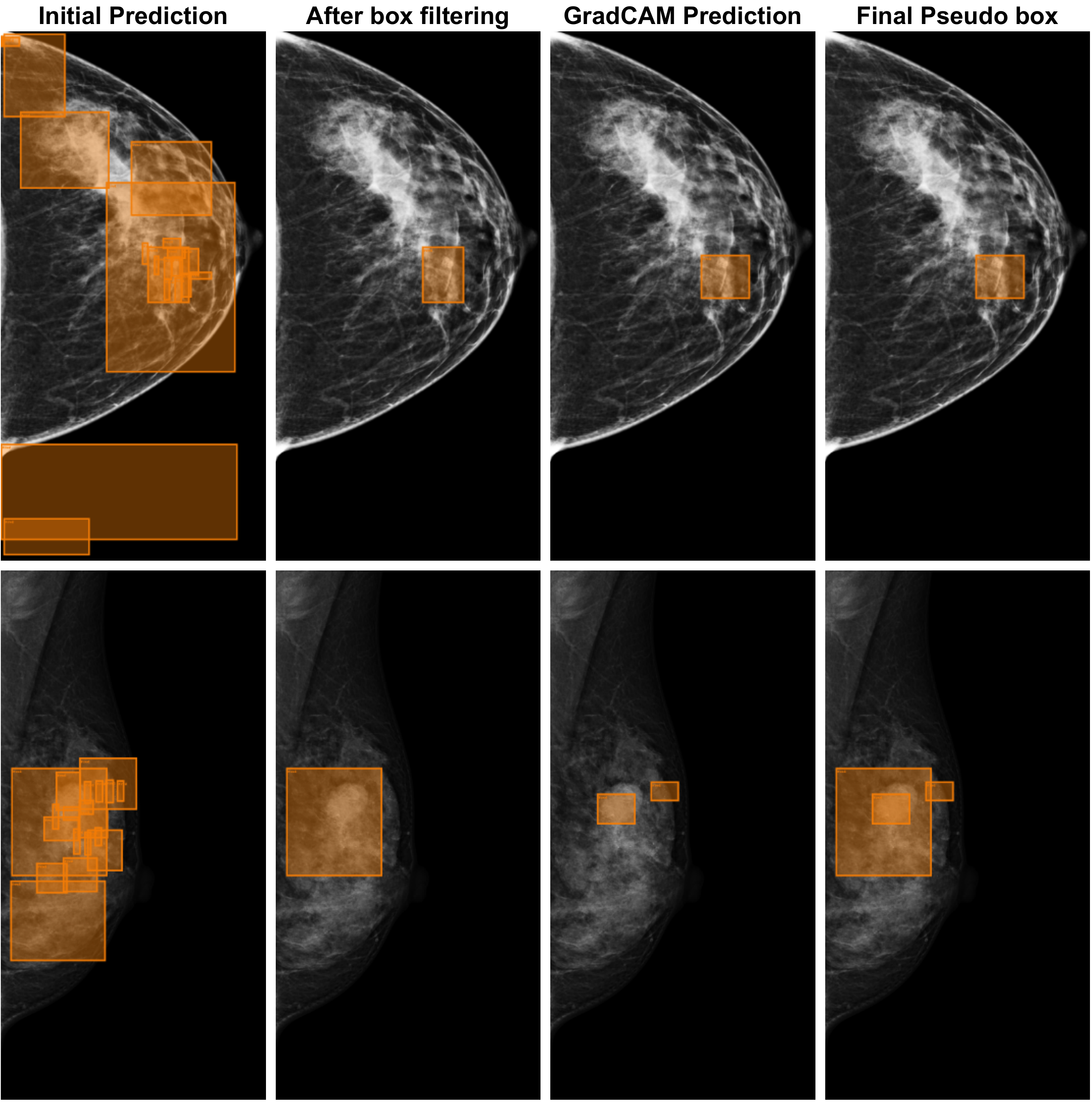}
    \caption{Visualisation of the pseudo-labelling by the teacher using a CC (top) and an MLO (bottom) view of a weakly-annotated sample. 
    The process starts with the initial predictions by the BRAIxDet teacher (first column), which is filtered to keep the detection with the highest score (second column). 
    Next, we show the GradCAM detections produced by BRAIxMVCCL~\citep{chen2022multi} (third column), and the final pseudo labels produced by the teacher to train the student with all GradCAM detections and the top BRAIxDet teacher detection (last column).}
    \label{fig:visual_result_pseudo_label}
\end{figure}

\begin{figure*}[ht]
    \centering
    \includegraphics[width=0.82\textwidth]{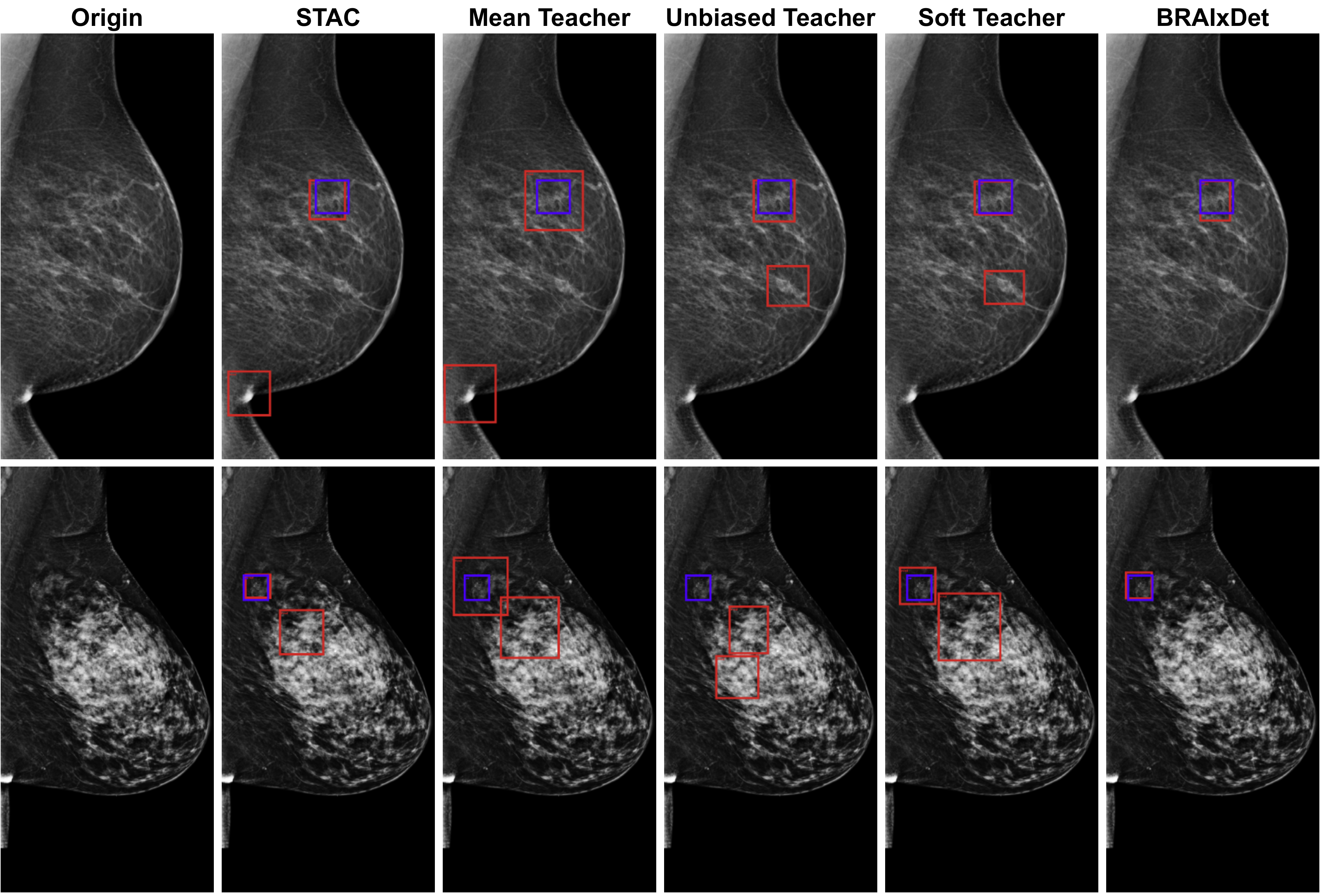}
    \caption{Visualisation of the results achieved by the most competitive methods and our BRAIxDet (column titles show each method's name) on two ADMANI testing images, where the \textcolor{blue}{blue} rectangles show lesion annotations and the \textcolor{red}{red} rectangles display the detections. 
    }
    \label{fig:visual_results_admani}
\end{figure*}

\begin{figure*}[!t]
    \centering
    \includegraphics[width=0.82\textwidth]{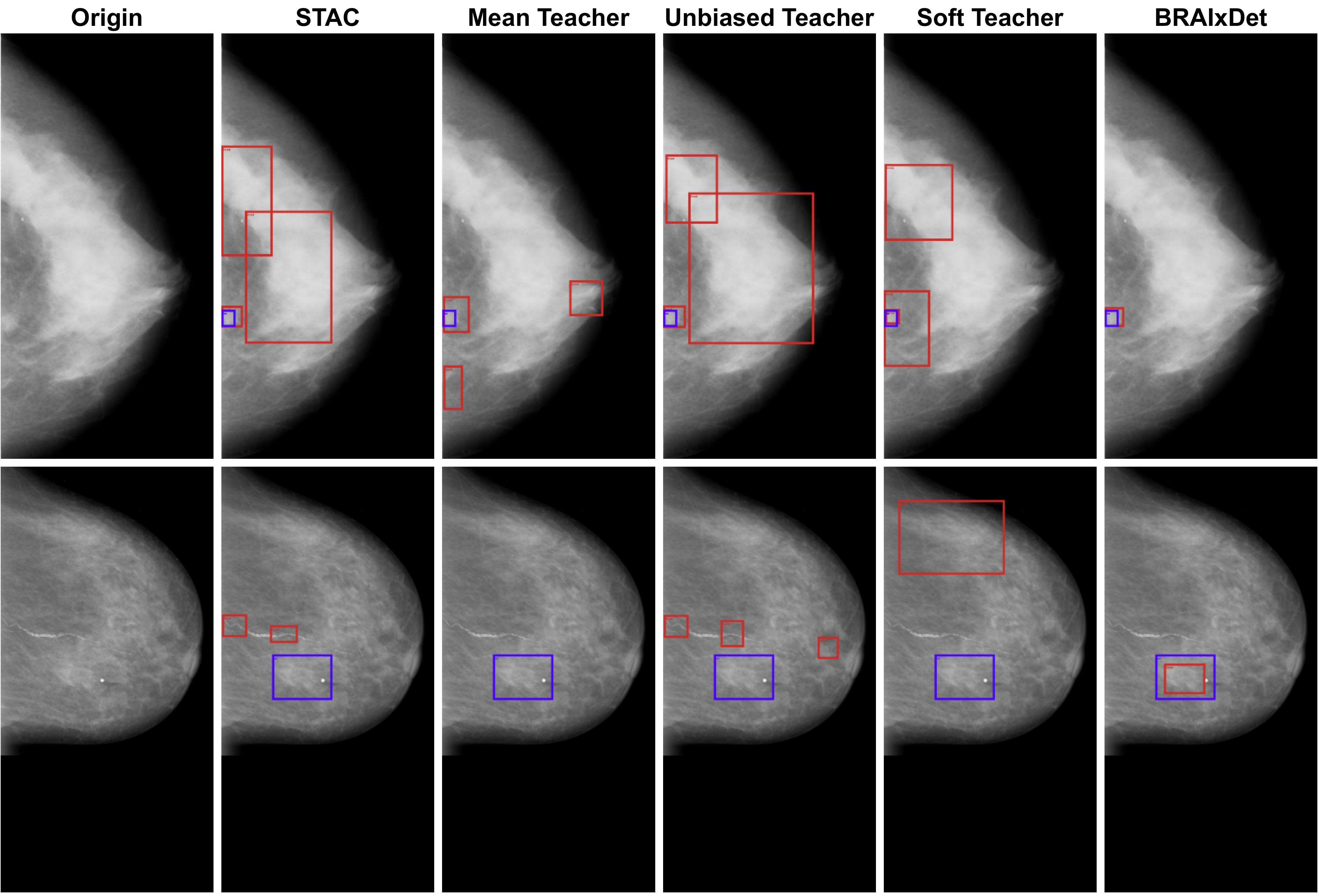} 
    \caption{Visualisation of the results achieved by the most competitive methods and our BRAIxDet (column titles show each method's name) on two CBIS-DDSM testing images, where the \textcolor{blue}{blue} rectangles show the lesion annotations, and the \textcolor{red}{red} rectangles display the detections. 
    }
    \label{fig:visual_results_ddsm}
\end{figure*}

\subsection{Implementation details}

We \textbf{pre-process each image} to remove text annotations and background noise outside the breast region, then we crop the area outside the breast region and pad the pre-processed images, such that their $H/W$ has the ratio $1536/768$ pixels. During data loading, we resize the input images to 1536 x 768 pixels and flip the images, so that the nipple is located on the right-hand side of the image. 
For the \textbf{pre-training}, we implement the classifier BRAIxMVCCL~\citep{chen2022multi} with the  EfficientNet-b0~\citep{tan2019efficientnet} backbone, initialized with ImageNet-trained~\citep{russakovsky2015imagenet} weights.
% pre-train
Our pre-training relies on the Adam optimiser~\citep{kingma2014adam} using a learning rate of 0.0001, weight decay of $10^{-6}$, batch size of 8 images and 20 epochs.
The \textbf{semi-supervised student-teacher learning} first updates the classifier BRAIxMVCCL~\citep{chen2022multi} into the proposed detector BRAIxDet using the Faster R-CNN~\citep{ren2015faster} backbone. We set the binarising threshold $\tau$ in Algorithm~\ref{alg:cam_to_box} to $0.5$.
% detection
We follow the default Faster R-CNN hyper-parameter setting from the Torchvision Library~\citep{paszke2019pytorch}, except for the NMS threshold that was reduced to $0.2$ and for the TP IoU detection for Roi\_head which was set to $0.2$. 
The optimization of BRAIxDet also relies on Adam optimiser~\citep{kingma2014adam} using a learning rate of 0.00005, weight decay of $10^{-5}$, batch size of 4 images and 20 epochs. Similarly to previous papers~\citep{tarvainen2017mean,liu2021unbiased}, we set $\lambda$, in the overall student loss of~\eqref{eq:overall_student_loss},
to 0.25, and $\alpha$ 
in the EMA to update the teacher's parameter in~\eqref{eq:teacher_ema_loss}
to 0.999.
For the pre-training and student-teacher learning stages, we use ReduceLROnPlateau to dynamically control the learning rate reduction based on the BCE loss during model validation, where the reduction factor is set to 0.1.

We compare our method to the following state-of-the-art (SOTA) approaches: 
Faster RCNN~\citep{ren2015faster}\footnote{\url{https://github.com/pytorch/vision}}, STAC~\citep{sohn2020simple}\footnote{\url{https://github.com/google-research/ssl_detection}}, MT~\citep{tarvainen2017mean}\footnote{\url{https://github.com/CuriousAI/mean-teacher}},
Unbiased Teacher~\citep{liu2021unbiased}\footnote{\url{https://github.com/facebookresearch/unbiased-teacher}.}, and 
Soft Teacher~\citep{xu2021end}\footnote{\url{https://github.com/microsoft/SoftTeacher}}.
All these competing methods' results are produced by running the code available from their official GitHub repository.
We additionally evaluate two malignant breast lesion detection methods: CVR-RCNN~\citep{ma2021cross} and MommiNet-V2~\citep{yang2021momminetv2} based on the Relation-Network~\citep{hu2018relation}\footnote{\url{https://github.com/heefe92/Relation_Networks-pytorch}}. Please note that we did not implement \textit{nipple detector} and \textit{image reg} modules in MommiNet-V2~\citep{yang2021momminetv2} due to a lack of details in their papers for reproduction.

We ran each experiment three times and formatted results in mean $\pm$ standard deviation.
These methods are implemented using Faster RCNN with EfficientNet-b0~\citep{tan2019efficientnet} backbone, pre-trained on the ADMANI dataset using the fully- and weakly-annotated training subsets, following the same setup as our BRAIxDet. 
After pre-training, while Faster RCNN is trained using only the fully-annotated subset, all other methods are trained with both the fully- and weakly-annotated training subsets.

All experiments are implemented with Pytorch~\citep{paszke2019pytorch} and conducted on an NVIDIA A40 GPU (48GB), where training takes about 30 hours on ADMANI and 8 hours on DDSM~\citep{web:ddsm}, and testing takes about 0.1s per image.
Given that the competing methods use Faster RCNN with the same backbone (EffientNet-b0) as ours and rely on pre-training, their training and testing running times are similar to ours.

\begin{figure*}[t]
    \centering
    \includegraphics[width=0.82\textwidth]{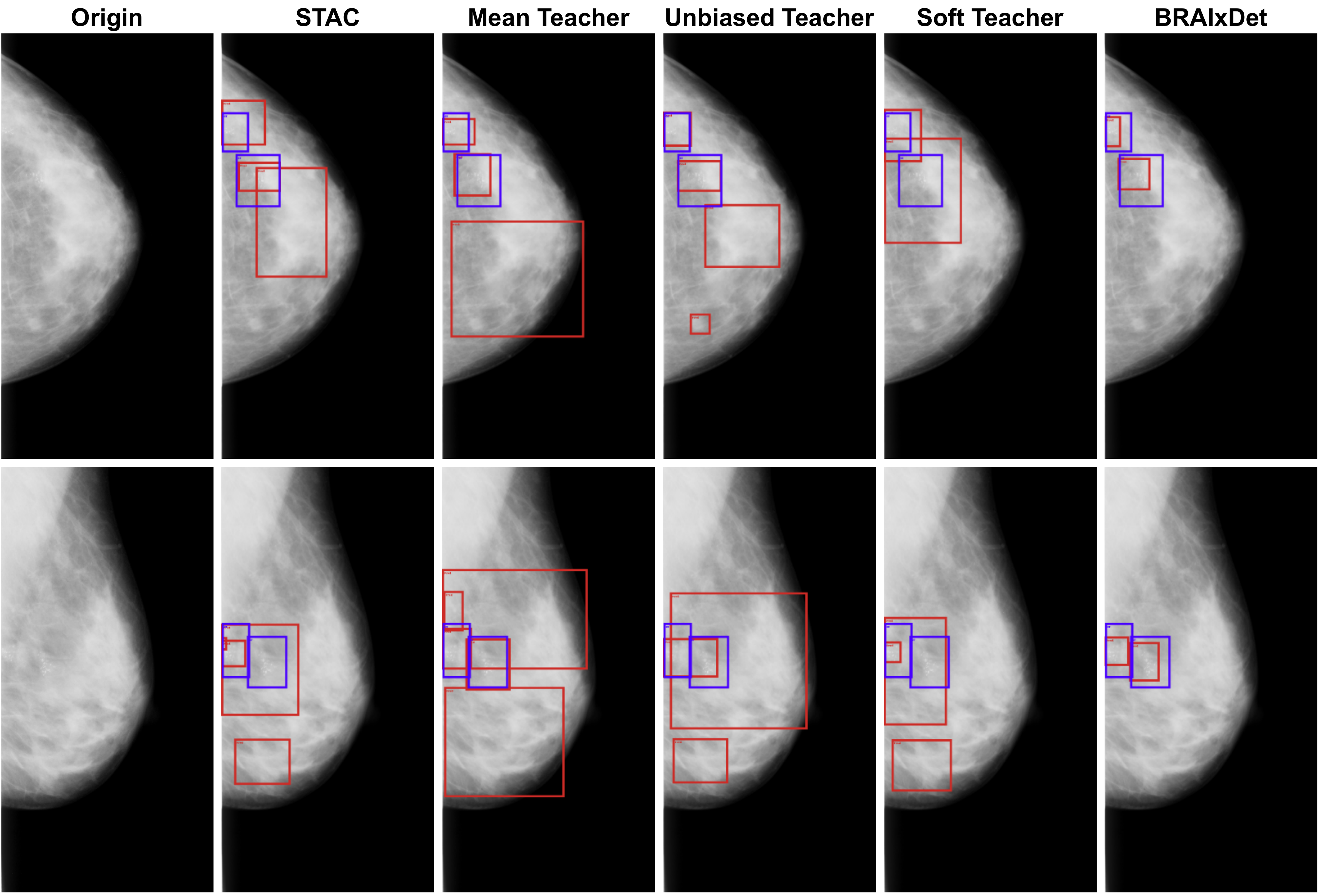}
    \caption{Visualisation of the cross-view results achieved by several methods and our BRAIxDet (column titles show each method's name) on CBIS-DDSM testing images containing two views of the same breast, where \textcolor{blue}{blue} rectangles show lesion annotations and \textcolor{red}{red} rectangles display the detections. 
    Note that BRAIxDet is the only method that correctly localises both lesions in the two views, showing the importance of the cross-view analysis.}
    \label{fig:cross_view_results}
\end{figure*}

\subsection{Results}

We first present the results produced by our BRAIxDet and competing approaches using the partially labelled protocol on ADMANI. 
Table~\ref{tab:admani_partial_result} shows the mAP and Recall @ 0.5 results, where BRAIxDet shows an mAP of \textcolor{r1}{$88.34$} for the ratio $1/16$, which is $6.5\%$ larger than the second best method, Soft Teacher.  For the ratio $1/2$, the mAP improvement is still large at around $4.5\%$. Similarly, for the Recall @ 0.5, BRAIxDet is around $7\%$ better than Soft Teacher (the second best method) for the ratio $1/16$, and when the ratio is $1/2$, the improvement is around $4\%$. 
The results from the partially labelled protocol on CBIS-DDSM on \textcolor{r1}{ Table~\ref{tab:ddsm_partial_result}} are similar to ADMANI. 
Indeed, our BRAIxDet shows consistently better results than other approaches for all training ratios. 

For instance, BRAIxDet presents an mAP of $53.99$ for ratio $1/16$, which is much larger than the second best, Unbiased Teacher, with a mAP of $36.18$. For ratio $1/2$, our mAP result is $5\%$ better than Soft Teacher (second best). 
Similar conclusions can be drawn from the results of the Recall @ 0.5.
Another interesting difference between the ADMANI and CBIS-DDSM results is the general lower performance on the CBIS-DDSM dataset. This can be explained by the lower quality of the images in the CBIS-DDSM dataset, as observed when comparing with Figures~\ref{fig:visual_results_admani} and~\ref{fig:visual_results_ddsm}.
It is worth noticing that in general, on both datasets, both BRAIxDet and competing SSL methods show better mAP and Recall @ 0.5 results than Faster RCNN, demonstrating the importance of using not only the fully-annotated cases, but also the weakly-annotated training subset.

The results produced by BRAIxDet and competing approaches using the fully labelled protocol on ADMANI using extra weakly-labelled data (the number
of extra weakly labelled images are defined inside the brackets)
are shown in 
Table~\ref{tab:admani_full_result}.
The mAP results confirm that BRAIxDet shows an improvement of $2.7\%$ with respect to the second best approach Soft Teacher. Similarly, 
BRAIxDet's Recall @ 0.5 results show an improvement of around $2.5\%$ over competing SSL methods.
Similarly to the partially labelled protocol from Table~\ref{tab:admani_partial_result},
BRAIxDet and competing SSL methods show better mAP and Recall @ 0.5 results than Faster RCNN.
We also present the testing performance results of malignant breast lesion detection models on supervised subsets of ADMANI~\citep{frazer2022admani} and DDSM~\citep{web:ddsm}, shown in Table~\ref{tab:supervised} (without using any extra weakly labelled images). We show that our approach also outperforms the previous breast lesion detection methods in the context of classic supervised learning. 
\textcolor{r1}{
We notice that there is a large gap between all models’ performance on ADMANI~\citep{frazer2022admani} and DDSM~\citep{web:ddsm}. For example, the mAP results on ADMANI testing images when training on the 1/16 partition (430 images) is 5.18\% better than the results on DDSM with 3/4 partition (780 images). The major reason for the discrepancy in performance is attributed to the data quality as the DDSM was completed in 1997, whereas the ADMANI dataset was collected between 2013 and 2019 and completed in 2021.
}

We performed a one-tailed t-test (between our BRAIxDet and the second best method) using the results from Tables~\ref{tab:admani_partial_result},~\ref{tab:ddsm_partial_result} and showed that we achieved $p$-values less than 0.05 for all partitions for both ADMANI and CBIS-DDSM datasets.  In Table~\ref{tab:admani_full_result}, we surpassed the second best method Soft Teacher~\citep{xu2021end} by 2.49\% and 2.47\% on the test set with $p$-values equal to $0.0007$ and $0.0001$, respectively for mAP and FROC measures. We conclude that our result is significantly better than previous methods for all experimental protocols.
\textcolor{r1}{
Moreover, the results in Fig.~\ref{tab:admani_partial_result} and Fig.~\ref{tab:ddsm_partial_result} reveal that for our BRAIxDet, a decrease in the amount of fully annotated data does not result in a significant decrease in model performance. This evidence indicates that our proposed method is more effective than other methods when applied to datasets with incomplete annotations at various levels of missing data.
}

We visualise the detection results by the most competitive methods and our proposed BRAIxDet model on ADMANI (Figure~\ref{fig:visual_results_admani}) and on CBIS-DDSM (Figure~\ref{fig:visual_results_ddsm}). In general, we can see that our method is more robust to false positive detections, while displaying more accurate true positive detections. We also show the cross-view detections on the CC and MLO mammograms from the same breast in Figure~\ref{fig:cross_view_results}. Notice how BRAIxDet is robust to false positives, and at the same time precise at detecting the true positives from both views. We believe that such cross-view accuracy is enabled in part by the cross-view analysis provided by BRAIxDet.

\begin{table}[t]
    \centering
    \caption{Comparison of different types of batch normalisation (BN) strategies for the student-teacher SSL stage.
    The first row shows the mAP and Recall @ 0.5 results using the usual approach, where the student updates its own BN statistics, but the teacher does not update the BN statistics from pre-training. Second row shows the competing approach~\citep{cai2021exponential} that updates the teacher's BN statistics with the EMA from the student's BN statistics.
    The last row shows our proposed approach based on freezing both the student and teacher BN statistics. Results are computed using the fully-labelled protocol on ADMANI dataset~\citep{frazer2022admani}.}
    \label{tab:bn}
    \resizebox{0.49\textwidth}{!}{
    \begin{tabular}{c||c||c}
    \toprule
    Methods	&	mAP	&	Recall @ 0.5	\\	\midrule
    Open BN~\citep{tarvainen2017mean} 	&	89.95 ± 0.52	&	92.71 ± 0.83	\\	
    EMA BN~\citep{cai2021exponential}	&	90.44 ± 0.63	&	93.05 ± 0.59	\\	
    Pre-estimated BN (ours)	&	\textbf{92.78 ± 0.78}	&	\textbf{95.18 ± 0.68}	\\	\bottomrule
    \end{tabular}
    }
\end{table}

\subsection{Ablation study}

In this ablation study, we first show a visual example of the steps in the production of a pseudo-label for weakly-annotated images by the teacher in Figure~\ref{fig:visual_result_pseudo_label}. The process starts with the initial prediction from the BRAIxDet teacher model, which tends to be inaccurate, particularly at the initial SSL training stages. 
This issue is addressed in part by keeping only the most confident detection, as shown in the column `After box filtering'. 
In addition, the BRAIxDet teacher's detection inaccuracies are also dealt with GradCAM predictions produced by BRAIxMVCCL~\citep{chen2022multi}, which are used with the most confident detection to produce the final pseudo-labelled detections. 

Table~\ref{tab:ablation} shows the impact of initial prediction results from GradCAM, the local co-occurrence module (LCM) from BRAIxMVCCL, the student-teacher SSL training, and the use of BRAIxMVCCL's GradCAM detections for the pseudo-labelling process.
All methods in this table first pre-train their EfficientNet-b0 classification backbone with the fully- and weakly-annotated training subsets.
The first row shows the original GradCAM results. The second row shows the Faster RCNN~\citep{ren2015faster} that is
trained using only the fully-annotated training subset. 
The third row displays a method that pre-trains the multi-view BRAIxMVCCL classifier~\citep{chen2022multi} and keeps the LCM for training the Faster RCNN, which shows that the multi-view analysis from LCM provides substantial improvement to the original Faster RCNN.
The fourth rows displays that our proposed student-teacher SSL training improves the original Faster RCNN results because such training allows the use of the fully- and weakly-annotated training subsets. 
When combining Faster RCNN, LCM, and the student-teacher SSL training (fifth row), the results are better than without SSL training or without LCM. 
Putting all BRAIxDet components together (sixth row), including BRAIxMVCCL's GradCAM detections, enables us to reach the best mAP and Recall @ 0.5 results.

Another important point that was investigated is the role of the proposed BN approach, which is presented in Table~\ref{tab:bn}. The results show that the proposed approach based on freezing both the student and teacher BN statistics is better than just updating the student's BN statistics, or updating both the student and teacher's BN statistics~\citep{cai2021exponential}.

% \textcolor{r2}{
% We further provide ablation studies on the hyperparameters of the model as shown in Fig.~\ref{fig:ablation_hyperp} including smoothing factor $\alpha$, heatmap threshold $\tau$ and weakly-supervised loss weight $\lambda$. 
% %
% In general, for $\alpha$, we note that increasing the smoothing factor make the EMA process less sensitive to inaccurate student weights, hence leading to an improved performance on both metrics, as shown in Fig.~\ref{fig:ablation_alpha}.
% %
% We set $\tau$ to 0.5 as it is a common default threshold and it represents a balance between the two classes. In our analysis, depicted in Fig.~\ref{fig:ablation_tau}, we conducted an ablation study on different values of $\tau$. The results reveal that $\tau=0.5$ exhibits a slight improvement over other threshold values.
% %
% Our empirical findings indicate that the best results are achieved when $\lambda=0.25$. We hypothesize that the predictions from the teacher model may exhibit instability, and thus assigning a smaller weight helps mitigate the impact of inaccurate gradient directions. 
% }

\textcolor{r2}{
In Fig.~\ref{fig:ablation_hyperp}, we provide further ablation studies on the hyper-parameters $\alpha$ for EMA decay, the threshold $\tau$ for binarising the GradCAM heatmap, and the weighting factor $\lambda$ that controls the contribution of the weakly-supervised loss. 
In general, we note that increasing the smoothing factor $\alpha$ makes the EMA process more robust to inaccurate student weights, leading to improved performance, as shown in Fig.~\ref{fig:ablation_alpha}. 
Also, the study of $\tau$ in Fig.~\ref{fig:ablation_tau} shows that $\tau=0.5$ exhibits a slight improvement over other threshold values.
Regarding $\lambda$ in Fig.~\ref{fig:ablation_lambda}, we notice that small values (e.g., $\lambda \in [0.1,0.5]$) lead to better performance, so we set it at $0.25$.
We hypothesize that the predictions from the teacher model may exhibit instability, and thus assigning a smaller weight helps mitigate the impact of inaccurate gradient directions. 
}

\begin{figure}[h]
    \subfigure[]{\includegraphics[width=0.24\linewidth]{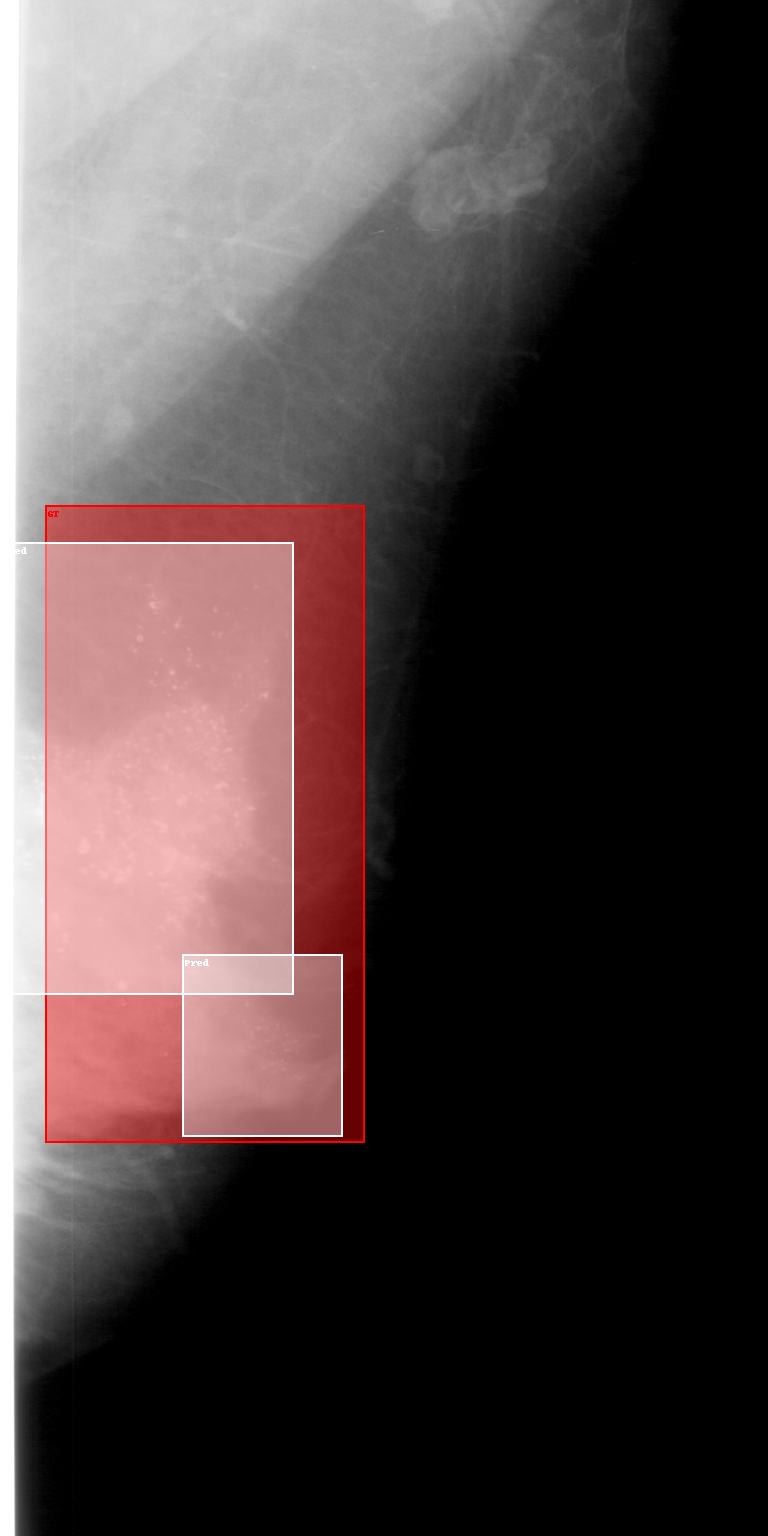}\label{fig:fail_multi_instance}}
    \subfigure[]{\includegraphics[width=0.24\linewidth]{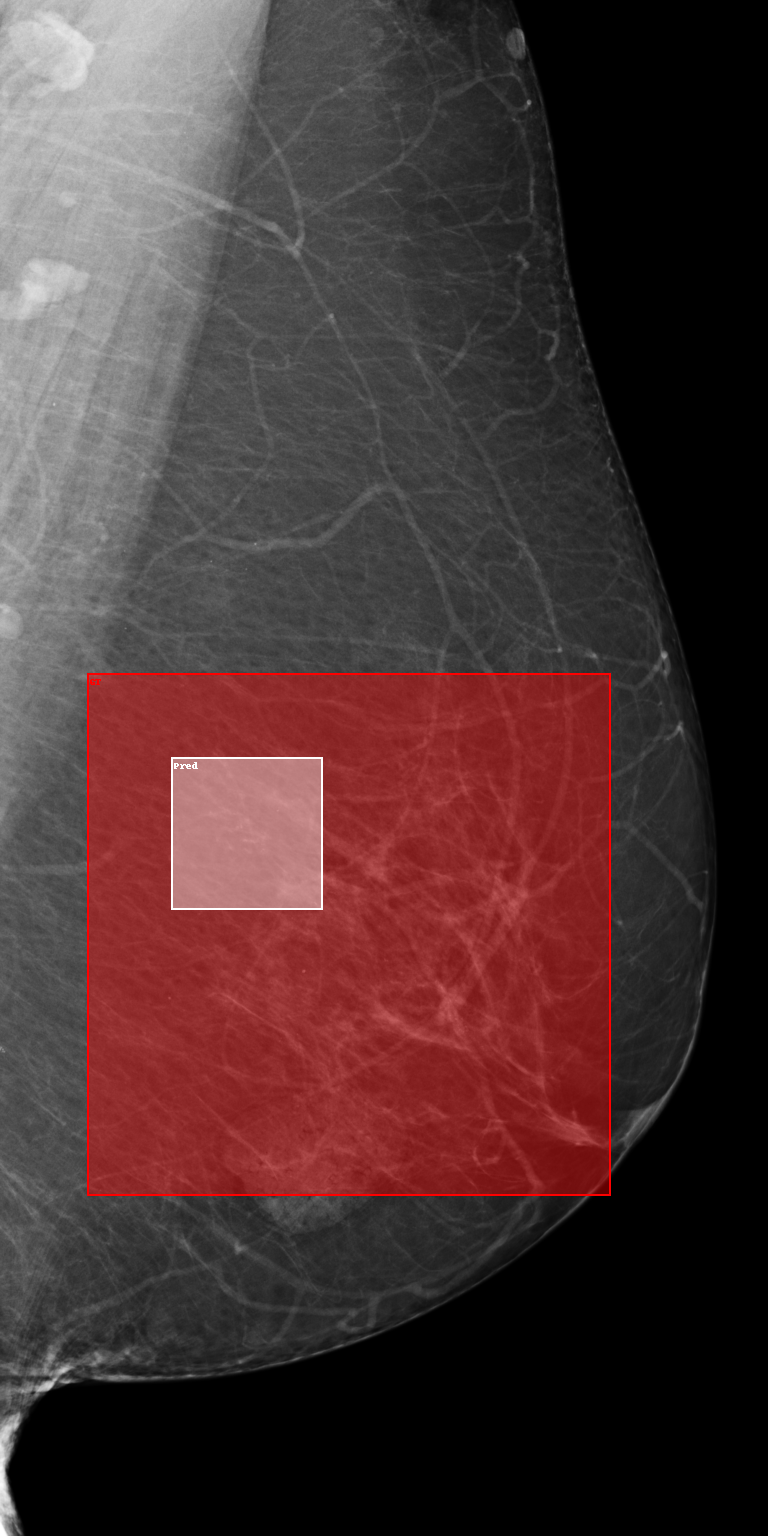}\label{fig:fail_overlarge}}
    \subfigure[]{\includegraphics[width=0.24\linewidth]{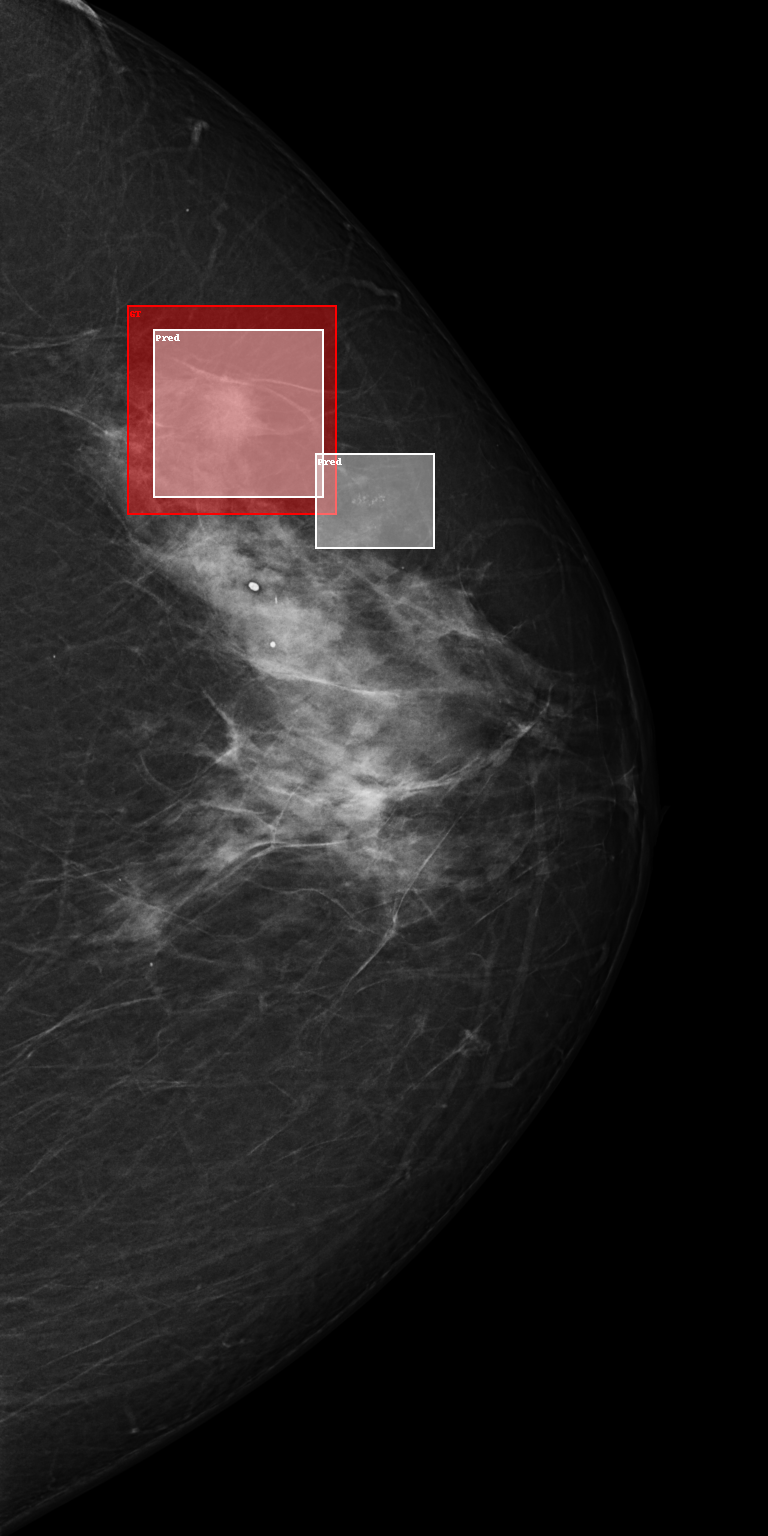}\label{fig:fail_benign}}
    \subfigure[]{\includegraphics[width=0.24\linewidth]{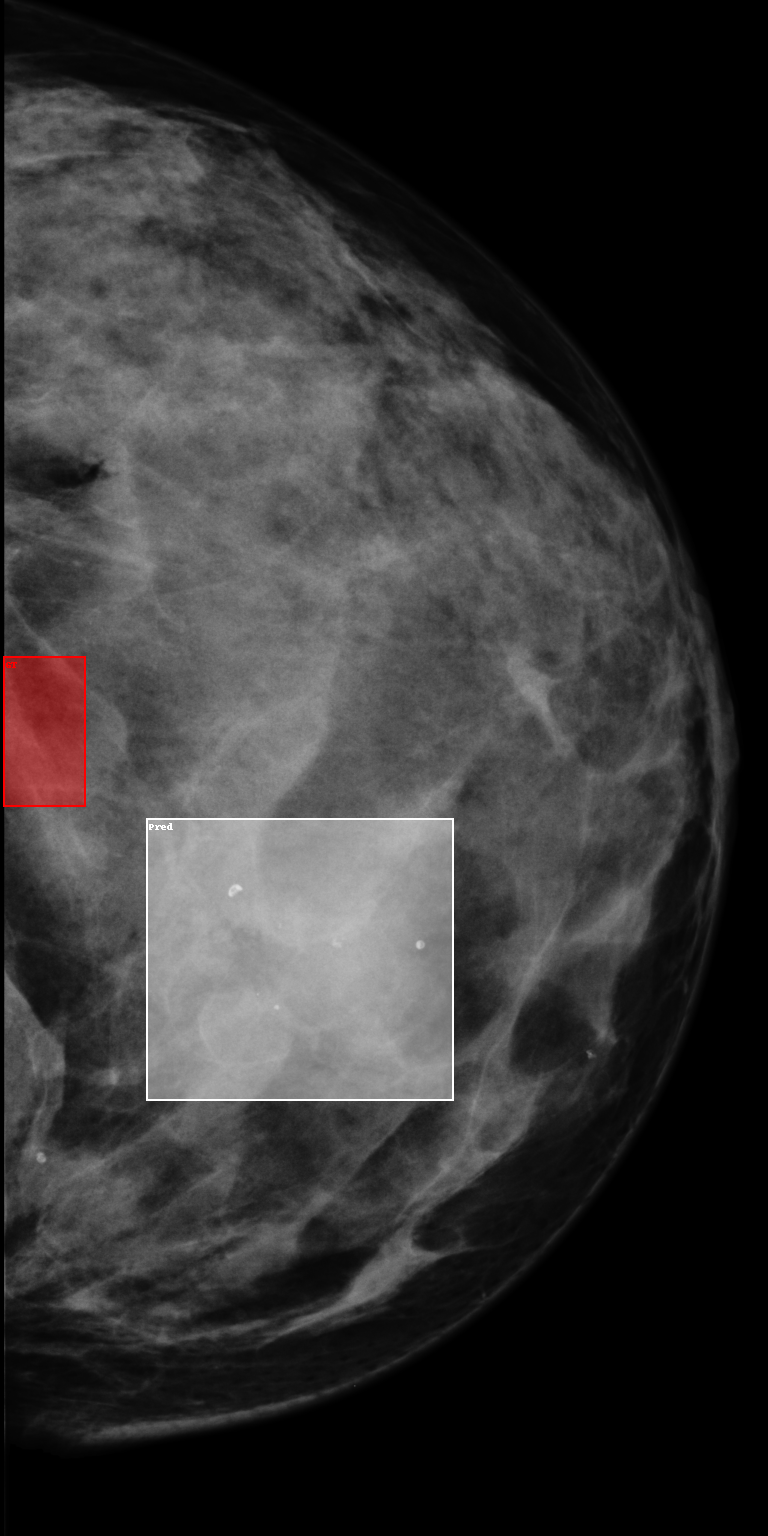}\label{fig:fail_unobserved}}
    \caption{Visualisations of four failure cases of BRAIxDet, where red boxes denote the manual labels, and white boxes represent the automatic detections from our system. The failure cases are the following: \textcolor{blue}{(a)} multiple detected regions inside a large ground-truth annotation;\textcolor{blue}{(b)} a small prediction within an excessively large annotation; \textcolor{blue}{(c)} falsely detected benign region (i.e., false positive); and \textcolor{blue}{(d)} miss-detected challenging malignant region (i.e., false positive and false negative).}
    \label{fig:failed_case}
\end{figure}

\subsection{Failure Cases} 
We show four failure cases produced by our system in Figure~\ref{fig:failed_case}. 
We notice that CBIS-DDSM~\citep{web:ddsm} and ADMANI~\citep{frazer2022admani} often have overly large annotations, as shown in Fig.~\ref{fig:failed_case}(\textcolor{blue}{a}) and Fig.~\ref{fig:failed_case}(\textcolor{blue}{b}). Additionally, our findings reveal that the system faces challenges in distinguishing between benign and malignant lesions, particularly when it comes to benign calcification, as illustrated in Figure~\ref{fig:failed_case}(\textcolor{blue}{c}). This is due to the absence of benign annotations in the ADMANI dataset~\citep{frazer2022admani}. In addition, we observed that the current model is challenged by lesions represented by subtle changes compared to normal breast tissue, as shown in Figure~\ref{fig:failed_case}(\textcolor{blue}{d}), which we believe is a significant challenge for any automated mammogram lesion detection system.
\subsection{Limitations.} A relevant limitation of our method is related to Grad-CAM, which is critical for the model, but can introduce some confirmation bias to the system as GradCAM only provides a coarse localization result that may include noisy labels. Also, localization methods like GradCAM will have difficulty identifying multiple overlapping objects as it is not designed to segment multiple independent instances. 
A limitation related to the annotations present in our datasets is that small lesions (e.g., micro-calcifications) can be detected as multiple bounding boxes, but the datasets ADMANI~\citep{frazer2022admani} and CBIS-DDSM~\citep{web:ddsm} tend to identify the areas that contain multiple micro-calcifications as a single instance. This discrepancy between the annotations and detections can lead to an artificial decrease in performance. 
A computational limitation is that the use of
weakly annotated datasets motivate us to introduce an additional classification network, thereby introducing an increase in computational costs for the system.
Another important limitation is that biases in the datasets (e.g., imaging manufacturers, population distribution) may impact our results.

\section{Conclusion}
In this paper, we proposed a new framework for training breast cancer lesion detectors from mammograms using real-world screening mammogram datasets.
The proposed framework contains a new problem setting and a new method.
The new setting is based on large-scale real-world screening mammogram datasets, which have a subset that is fully annotated and another subset that is weakly annotated
with just the global image classification and without lesion localisation -- we call this setting malignant breast lesion detection with incomplete annotations.
Our solution to this setting is a new method with the following two stages: 1) pre-training a multi-view mammogram classifier with weak supervision from the whole dataset, and 2) extend the trained classifier to become a multi-view detector that is trained with semi-supervised student-teacher learning, where the training set contains fully and weakly-annotated mammograms.
We provide extensive detection results on two real-world screening mammogram datasets containing fully and weakly annotated mammograms, which show that our proposed approach has SOTA results in the detection of malignant breast lesions with incomplete annotations. In the future, we aim to further improve model performance by
incorporating additional multi-modal information (i.e., radiology reports, risk factors or ultrasound) to calibrate the confidence of the pseudo-label.

\section*{Acknowledgments}

This work was supported by funding from the Australian Government under the Medical Research Future Fund - Grant MRFAI000090 for the Transforming Breast Cancer Screening with Artificial Intelligence (BRAIx) Project. G. Carneiro acknowledges the support by the Australian Research Council through grants DP180103232 and FT190100525.

%%Harvard
\bibliographystyle{model2-names.bst}\biboptions{authoryear}
\bibliography{refs}

% \section*{Supplementary Material}

% Supplementary material that may be helpful in the review process should
% be prepared and provided as a separate electronic file. That file can
% then be transformed into PDF format and submitted along with the
% manuscript and graphic files to the appropriate editorial office.

\end{document}